\documentclass{article}
\usepackage{xcolor}         

\usepackage{PRIMEarxiv}

\usepackage[utf8]{inputenc} 
\usepackage[T1]{fontenc}    
\usepackage{hyperref}       
\usepackage{url}            
\usepackage{booktabs}       
\usepackage{amsfonts}       
\usepackage{nicefrac}       
\usepackage{microtype}      
\usepackage{bm}             
\usepackage{siunitx}        
\usepackage{longtable}      
\usepackage{lipsum}
\usepackage{fancyhdr}       
\usepackage{graphicx}       
\graphicspath{{media/}}     
\usepackage{xcolor}         
\usepackage{adjustbox}
\usepackage{threeparttable}

\usepackage{lineno}
\usepackage{amsthm}
\usepackage{amsmath}
\usepackage{amssymb}
\usepackage{algorithm}
\usepackage{algorithmic}
\usepackage{subcaption}
\usepackage{tikz}
\usepackage{booktabs}
\usepackage{multirow}
\usepackage{tabularx}

\modulolinenumbers[5]

\usepackage{amsthm}

\title{Pretrained Approximators for Low-Thrust Trajectory Cost and Reachability}

\author{
  \textbf{Zhong Zhang}\textsuperscript{*} \\
  Politecnico di Milano \\
  \texttt{zhong.zhang@polimi.it}
  \And
  \textbf{Giacomo Acciarini}\textsuperscript{*} \\
  European Space Agency\\
  \texttt{giacomo.acciarini@esa.int}
  \And
  \textbf{Dario Izzo}\textsuperscript{*} \\
  European Space Agency \\
  \texttt{dario.izzo@esa.int}
  \And
  \textbf{Hexi Baoyin}\\
  Tsinghua University\\
  \texttt{baoyin@tsinghua.edu.cn}
  \And
  \textbf{Francesco Topputo}\\
  Politecnico di Milano \\
  \texttt{francesco.topputo@polimi.it}
}
\begin{document}

\maketitle

\renewcommand{\thefootnote}{\fnsymbol{footnote}}
\footnotetext[1]{These authors contributed equally to this work.}

\newtheorem{proposition}{Proposition}
\newtheorem{theorem}{Theorem}
\newtheorem{lemma}{Lemma}
\newtheorem{corollary}{Corollary}

\begin{abstract}
Low-thrust trajectory design relies heavily on repeated evaluations of fuel
consumption and transfer feasibility, which require expensive optimal control solutions. In this work, we show these quantities can be accurately approximated by machine learning surrogates, 
enabling fast and scalable evaluation
across a wide range of scenarios.
By increasing both dataset size and model capacity, we observe that low-thrust trajectory optimization follows a scaling law, with performance improving linearly with the logarithm of training data and network parameters, and no evidence of predicted saturation within the explored regime. Guided by this empirical observation, we construct a large-scale dataset using the proposed homotopy-ray strategy tailored to mission design requirements.
A key is the introduction of a self-similar transformation, which allows generalization across semi-major axes, inclinations, and central bodies avoiding retraining.
As a result, the same neural approximator can be applied to diverse orbital environments and mission classes. 
The proposed models accurately predict optimal fuel consumption and minimum transfer time for single- and multi-revolution transfers. 
Their performance and generalization are demonstrated on a public dataset, a multi-asteroid flyby problem from the Global Trajectory Optimization Competition, and an asteroid rendezvous mission design. 
The models and datasets are released as open-source to support the space community.
\end{abstract}

\section{Introduction}
Low-thrust electric propulsion has become a key enabling technology for modern
space missions, offering substantially higher specific impulse and mission
flexibility compared to conventional chemical propulsion. Its successful
application in missions such as Deep Space~1~\cite{raymanMissionDesignDeep1999},
Hayabusa~\cite{kawaguchiHayabusaItsTechnology2008}, BepiColombo~\cite{benkhoffBepiColomboComprehensiveExploration2010},
and Psyche~\cite{oh2019development} has demonstrated its potential for ambitious
scientific and exploration objectives. At the same time, low-thrust propulsion
fundamentally changes the nature of trajectory design: optimal transfers are no
longer defined by a small number of impulsive maneuvers, but by continuous
control laws obtained from solving nonlinear optimal control problems\cite{zhangSustainableAsteroidMining2025}.

While high-fidelity optimal control methods are well established, their
computational cost remains a major bottleneck in practical mission design.
During early-phase analysis and global optimization, mission designers often
need to evaluate millions of candidate transfers under varying boundary
conditions, propulsion parameters, and mission constraints~\cite{zhangGlobalTrajectoryOptimization2024}. Solving a full
low-thrust optimal control problem for each candidate is generally
computationally prohibitive, motivating the widespread use of approximate
methods for estimating key performance indicators such as fuel consumption,
transfer duration, and reachability.
As physicist Lev Landau noted, “the most important part of doing physics is the
knowledge of approximation,” a principle that equally applies to this case. 

A wide range of approximation techniques have been developed for this purpose.
Analytical and semi-analytical approaches, including Edelbaum-type
approximations~\cite{edelbaum1961propulsion}, Lambert-based surrogates~\cite{izzoRevisitingLambertsProblem2015,woollands2015new,russellCompleteLambertSolver2022},
shape-based methods~\cite{wallShapeBasedApproachLowThrust2009,wuAnalyticalShapingMethod2022},
and simplified dynamical models~\cite{gurfil2004analysis,casalino2014approximate,hennesFastApproximatorsOptimal2016,shen2021simple}, offer excellent computational
efficiency, but are typically restricted to specific dynamical regimes or
mission geometries. Database-driven approaches can achieve higher accuracy over
narrowly defined problem classes~\cite{petropoulos_anastassios_2018_1139152,zhangGTOC11Results2023a},
yet their performance strongly depends on data coverage and often degrades when
applied outside the original data domain.

In recent years, machine-learning-based surrogate models have emerged as a
promising alternative, combining fast evaluation with improved expressive
power. Neural networks, as universal function approximators~\cite{parkUniversalApproximationUsing1991},
have been successfully applied to a variety of astrodynamics problems, including orbit uncertainty, periodic
orbit generation in multi-body dynamics, on-board guidance, and autonomous
navigation\cite{harlNeuralNetworkBased2013,wilsonGenerationClassificationCritical2024,federiciImageBasedMetaReinforcementLearning2022,izzoRealTimeGuidanceLowThrust2021,izzoNeuralRepresentationTime2023,pugliattiOnboardStateEstimation2024, ACCIARINI2025694}.

For low-thrust trajectory optimization, these methods may be used either to estimate trajectory-level performance quantities or to approximate directly optimal control profiles. This work focuses on the former setting, namely the fast prediction of fuel consumption and reachability. The detailed terminology and representative studies are reviewed in Sec.~\ref{sec:relatedwork}.

Despite encouraging results, current machine learning-based approaches face
several limitations that hinder their broader adoption in mission design.
First, most existing models are trained for narrowly defined mission scenarios,
with fixed propulsion characteristics or limited orbital regimes, requiring
data regeneration and retraining when applied to new scenarios. Second, the lack of
publicly available datasets and models makes independent validation
and comparison difficult, reducing confidence in neural surrogate methods among
practitioners.

The objective of this work is to address these limitations by developing a
general-purpose neural approximator for low-thrust trajectory
evaluation that can be applied across a wide range of mission scenarios without
retraining. To this end, we make four main contributions. First, we provide
evidence that low-thrust trajectory approximation exhibits a scaling law:
increasing both dataset size and model capacity leads to systematic prediction performance
improvements without observable saturation within the explored range. Second,
motivated by this observation, we construct a large-scale dataset using a proposed
homotopy-ray data generation method specifically designed to match typical
mission design requirements. Third, we introduce a self-similar transformation
of the problem variables that enables a single pretrained model to generalize
across different semi-major axes, inclinations, and central bodies. Finally, we
release both the trained models and the dataset as open-source resources.

Compared with existing reports, the proposed models achieve better accuracy in predicting optimal fuel
consumption and minimum transfer time. Their performance and generalization capabilities are demonstrated on a public
dataset, the 4th Global Trajectory
Optimization Competition problem, and a low-thrust asteroid rendezvous scenario.

The remainder of this paper is organized as follows. Section~\ref{sec:relatedwork}
reviews related work on neural approximators for low-thrust trajectory design.
Section~\ref{sec:DataGeneration} describes the optimal control formulation and
the homotopy-ray data generation method. Section~\ref{sec:input_output_analysis}
introduces the self-similar transformation and the selection of network inputs.
Sections~\ref{sec:NNTraining} and~\ref{sec:NNTraining_MultiRev} detail the neural
network architectures and training strategies for single- and multi-revolution
transfers, respectively. Section~\ref{sec:results} presents numerical results
and benchmark studies, and Section~\ref{sec:conclusion} concludes the paper.

\section{Related Work}
\label{sec:relatedwork}

Neural-network-based methods for low-thrust trajectory design generally address
two complementary problems~\cite{izzo2024optimality}. The first concerns the prediction of
trajectory-level performance indicators, such as fuel consumption, transfer
time, or reachability. The second focuses on the direct approximation of optimal
control laws or costates, enabling real-time guidance. Following the terminology
introduced in~\cite{izzoRealTimeGuidanceLowThrust2021}, these two classes of
models are referred to as \emph{Value Networks} and \emph{Policy Networks},
respectively. Value Networks are primarily intended for mission design and
analysis, where fast evaluation of large numbers of candidate transfers is
required, whereas Policy Networks are designed for on-board guidance and
control. This section reviews representative work in both categories, with an
emphasis on Value Networks, which are the focus of this study.

\subsection{Value Network}
\label{sec:valuenetwork}

Value Networks aim to replace computationally intensive low-thrust trajectory
optimization with fast surrogate models that approximate the mapping from
mission parameters to optimal performance metrics. Their main application lies
in preliminary mission design and global optimization, where millions of
trajectory evaluations may be required~\cite{OzakiNaoya}.

The early work demonstrating the application of machine learning to approximate fuel-optimal low-thrust transfers was presented by Mereta et al.~\cite{meretaMachineLearningOptimal2017}. Subsequently, Zhu et al.~\cite{zhuFastEvaluationLowThrust2019} and Li et al.~\cite{liDeepNetworksApproximators2020} further investigated this direction.
A common formulation involves the use of two separate networks: a classifier to
determine transfer feasibility and a regressor to estimate fuel
consumption~\cite{zhuFastEvaluationLowThrust2019,XIE2021107002}. Li et
al.~\cite{liDeepNetworksApproximators2020} further extended this approach to
impulsive maneuver problems, highlighting the versatility of neural
approximators across propulsion regimes.

For minimum-transfer-time problems, Guo et
al.~\cite{guoDNNEstimationLowthrust2023} addressed the challenge of data sparsity
near the reachability boundary by introducing enhanced sampling strategies and
weighted loss functions, significantly improving prediction accuracy for
asteroid rendezvous missions. Mughal et
al.~\cite{MUGHAL2022200092} similarly employed deep neural networks to predict
electric-propulsion transfer times for LEO-to-GEO missions, achieving rapid and
accurate estimates suitable for preliminary design studies.

Several studies have further demonstrated the effectiveness of Value Networks in capturing nonlinear optimal-control structures that are challenging to model analytically. 
The neural representation of optimal guidance problems was formalized by Izzo and Origer~\cite{izzoNeuralRepresentationTime2023}. Acciarini et al.~\cite{acciariniComputingLowthrustTransfers2024a} investigated neural approximators for low-thrust transfer problems and benchmarked their performance against analytical solutions.
Extensions to more complex
dynamical environments, including multi-target solar-sailing
missions~\cite{SONG201928} and proximity operations under
perturbations~\cite{math10142489}, further demonstrate the flexibility and broad
applicability of Value Networks in astrodynamics.

\subsection{Policy Network}

Policy Networks address a different problem: the direct approximation of
optimal control laws or costates for real-time guidance. The work
by Sánchez-Sánchez and Izzo~\cite{doi:10.2514/1.G002357,7850105} introduced the
use of supervised learning, or behavioral cloning, to train neural networks
that generate near-optimal low-thrust guidance in real time. These architectures,
now commonly referred to as G\&CNETs, marked a significant step toward the use
of neural networks in mission-critical on-board systems.

Subsequent studies investigated the theoretical properties of such systems.
Izzo et al.~\cite{izzo2020stability} demonstrated that stability guarantees can
be established for G\&CNET-based guidance laws, providing a foundation for their
use in safety-critical applications. This work has motivated a wide range of
applications, including lunar and Mars landings, irregular asteroid landings,
interplanetary transfers, solar sailing, and proximity
operations~\cite{cheng2020real,cheng2018real,gaglio2025machine,tailor2019learning,federici2021deep}.

To improve robustness under model uncertainties and environmental disturbances,
researchers have integrated deep reinforcement learning and meta-learning
techniques into neural guidance frameworks. Gaudet et
al.~\cite{gaudet2020deep,gaudet2020adaptive} showed that these approaches enhance
policy robustness and enable effective sim-to-real transfer in the presence of
unmodeled dynamics. Evans et al.~\cite{EVANS202517} investigated the use of
differential algebra to refine neural guidance laws, while Holt et
al.~\cite{holt2025reinforcement} provided comparative analyses of reinforcement
learning and supervised learning approaches for guidance applications.

Recent research has sought to connect neural control with classical stability and safety theory. Approaches based on Lyapunov functions provide formal guarantees of stability and constraint satisfaction~\cite{holt2025reinforcement,holt2024reinforced,sun2025guaranteed}, while the event-transition tensor framework introduced by Izzo et al.~\cite{izzo2025highorderexpansionneuralordinary} offers a systematic tool for analyzing the behavior of G\&CNETs when deployed as controllers for uncertain dynamical systems.

\section{Data Generation}
\label{sec:DataGeneration}

Constructing large-scale, mission-oriented low-thrust trajectory datasets
requires both accurate trajectory solutions and efficient coverage of the
engineering-relevant state space. In this work, we adopt an indirect optimal
control method to compute individual trajectories, and combine it with a
continuation strategy, referred to as the \emph{Homotopy Ray Method}, to generate
a diverse set of transfer solutions. 

Two separate datasets are constructed to train dedicated neural networks for
fuel-optimal and time-optimal problems, respectively. Reachability naturally
arises from the time-optimal formulation: a transfer is unreachable if the
specified transfer time is shorter than the minimum, and reachable only once
this lower bound is exceeded. Trajectories at the boundary of reachability
correspond to the time-optimal solution~\cite{zhuFastEvaluationLowThrust2019}. Framing reachability prediction as a time-optimal control problem
thus provides a continuous and physically meaningful measure, offering richer
information than a simple binary classification of reachable versus
unreachable transfers.

\subsection{Indirect Method for Optimal Control}
This subsection describes the indirect method used to solve fuel-optimal and
time-optimal low-thrust optimal control problems. In both cases, Pontryagin's
minimum principle is applied by introducing the costate vector to transform the
original optimal control problem into a two-point boundary value problem
(TPBVP), which is then solved with a shooting method. In all the problem formulations, we assume that the maximum thrust $T_{\max}$ and specific impulse $I_{\rm sp}$ are constants (i.e., independent of the distance between spacecraft to the central body).

\subsubsection{Problem Formulations}
\label{sec:problem_formulations}

For the fuel-optimal problem, the objective is to minimize the propellant
consumption:
\begin{align}
  \min_{\boldsymbol{u}} \quad & J_{\rm fuel}
  \label{eq:fuel_obj}                                                                                                                      \\
  \text{s.t.} \quad
                              & \dot{\boldsymbol{x}}(t) = \boldsymbol{f}(\boldsymbol{x}) + \boldsymbol{g}(\boldsymbol{x,u}),               \\
                              & \dot{m}(t) = -\frac{T_{\max} \, \|\boldsymbol{u}(t)\|}{I_{\rm sp} g_0}, \quad \|\boldsymbol{u}(t)\| \leq 1 \\
                              & \boldsymbol{x}(t_0) = \boldsymbol{x}_0, \quad m(t_0) = m_0,                                                \\
                              & \boldsymbol{x}(t_f) = \boldsymbol{x}_f.\label{eq:fuel_final_state}
\end{align}

For the time-optimal problem, the goal is to minimize the transfer time:
\begin{align}
  \min_{\boldsymbol{u}, t_f} \quad & J_{\rm time}
  , \label{eq:time_obj}                                                                                                                         \\
  \text{s.t.} \quad
                                   & \dot{\boldsymbol{x}}(t) = \boldsymbol{f}(\boldsymbol{x}) + \boldsymbol{g}(\boldsymbol{x,u}),               \\
                                   & \dot{m}(t) = -\frac{T_{\max} \, \|\boldsymbol{u}(t)\|}{I_{\rm sp} g_0}, \quad \|\boldsymbol{u}(t)\| \leq 1 \\
                                   & \boldsymbol{x}(t_0) = \boldsymbol{x}_0, \quad m(t_0) = m_0,                                                \\
                                   & \boldsymbol{x}(t_f) = \boldsymbol{x}_f(t_f), \label{eq:time_final_state}
\end{align}
where the 
$\boldsymbol{f}$ denotes the gravitational dynamics, while $\boldsymbol{g}$ represents the thrust-induced dynamics. 
Before presenting the indirect method for solving optimal control problems, we first introduce four techniques that we have been employing to facilitate their solution:
\begin{enumerate}
  \item Instead of using Cartesian coordinates $(\boldsymbol{r}, \boldsymbol{v})$ or
        classical orbital elements $(a, e, i, \Omega, \omega, \theta)$, the state vector
        $\boldsymbol{x}$ is represented using modified equinoctial elements
        (MEE)\cite{walkerSetModifiedEquinoctial1985}, which are non-singular and
        well-suited for solving majority of low-thrust trajectory optimization
        problems\cite{junkinsExplorationAlternativeState2019}.
        \begin{equation}
          \begin{array}{ccc}
            p = a (1 - e^2),            & f = e \cos (\omega + \Omega), & g = e \sin (\omega + \Omega),   \\
            h = \tan (i/2) \cos \Omega, & k = \tan (i/2) \sin \Omega,   & L = \omega + \Omega + \theta,
          \end{array}
        \end{equation}
        Defining the spatial state vector as $\bm{x} = [p,\,f,\,g,\,h,\,k,\,L]^{\mathrm{T}}$, the dynamics are expressed as
        \begin{equation}
          \label{eq:indirect_dynamics}
          \dot{\bm{x}} = \bm{D}(\bm{x}) + \frac{T_{\max}}{m}  \bm{M}(\bm{x})\,\bm{\alpha} u
        \end{equation}
        where $\bm{D}(\bm{x})$ denotes the Keplerian drift vector field, and
        $\bm{M}$ is the transformation matrix linking the control to the MEE rates.
        Detailed matrix expressions for $\bm{M}$ and $\bm{D}$ can be found in supplementary material. Readers interested in the various formulations of these expressions may refer to the works of~\cite{izzoRealTimeGuidanceLowThrust2021,junkinsExplorationAlternativeState2019,gaoLowThrustInterplanetaryOrbit2004} for further details.

  \item To address the discontinuity of bang-bang control in fuel-optimal problems, a
        logarithmic homotopy method is applied to smooth the control profile, thereby
        improving convergence and solution
        efficiency~\cite{bertrandNewSmoothingTechniques2002}.
  \item To facilitate the initialization of the costate vector, a normalization
        technique is employed by introducing a scalar multiplier $\lambda_0$, such that
        the magnitude of the initial costate vector $\boldsymbol{\lambda}_0$ is set to
        1. This does not alter the nature of the optimal control problem, but improves
        the ability to estimate reasonable initial guesses for the costate values
        \cite{jiangPracticalTechniquesLowThrust2012a}. Therefore, the fuel-optimal cost
        function is reformulated as
        \begin{equation}
          \label{eq:indirect_log_homotopy}
          J_{\rm fuel} =       \lambda_0 \int_{t_0}^{t_f} L_{\rm fuel}(\bm{x}, \bm{u})\mathrm{d} t =
          \lambda_0 \int_{t_0}^{t_f} \frac{T_{\max }}{I_{\mathrm{sp}} g_0}  \{u-\varepsilon \ln [u(1-u)]\} \mathrm{d} t
        \end{equation}
        with $\epsilon = 1\times 10^{-5}$ in all generation process, which effectively smooths the control profile and maintains the high fidelity~\cite{bertrandNewSmoothingTechniques2002}. The time-optimal cost function is similarly defined as
        \begin{equation}
          \label{eq:indirect_log_homotopy_time}
          J_{\rm time} =
          \lambda_0 \int_{t_0}^{t_f} 1 \mathrm{d} t
        \end{equation}

  \item In the time-optimal problem, the final states are represented as functions of
        $t_f$ to enhance feasibility and are enforced as equality constraints in
        Eq.~\eqref{eq:time_final_state}, in contrast to the fixed terminal states in
        the fuel-optimal problem (Eq.~\eqref{eq:fuel_final_state}). 
        Since the target state evolves with time, the objective is to achieve rendezvous with the moving target in minimum time rather than reach a fixed terminal state.
\end{enumerate}

\subsubsection{Indirect method and TPBVP Formulation}
\label{sec:indirect_method_and_tpbvp_formulation}

For the fuel-optimal problem, referring to the costate vector
$\bm{\lambda}_x(t), {\lambda}_m(t)$ and scalar multiplier $\lambda_0$, the
Hamiltonian is constructed as
\begin{equation}
  H_{\rm fuel} = \bm{\lambda}_x^{\mathrm{T}} \dot{\bm{x}} + \bm{\lambda}_m \dot{m} + \lambda_0  L_{\rm fuel}(\bm{x}, \bm{u}) =
  \frac{T_{\max}}{m} \bm{\lambda}_x^{\mathrm{T}} \bm{M}\,\bm{\alpha} u + \bm{\lambda}_x^{\mathrm{T}} \bm{D} - \lambda_m \frac{T_{\max}}{I_{\rm sp} g_0}u + \lambda_0 \frac{T_{\max }}{I_{\mathrm{sp}} g_0}  \{u-\varepsilon \ln [u(1-u)]\}.
\end{equation}

The costate dynamics are then given by
\begin{equation}
  \label{eq:indirect_costate}
  \begin{aligned}
    \dot{\bm{\lambda}}_x
    & = -\frac{\partial H}{\partial \bm{x}}
      = -\left(
      \frac{T_{\max}}{m}
      \left(\frac{\partial \bm{M}}{\partial \bm{x}}\,\bm{\alpha}\right)^{\mathrm{T}}
      \bm{\lambda}_x\, u
      + \left(\frac{\partial \bm{D}}{\partial \bm{x}}\right)^{\mathrm{T}}
      \bm{\lambda}_x
      \right), \\
    \dot{{\lambda}}_m
    & = -\frac{\partial H}{\partial {m}}
      = \frac{T_{\max}}{m^2}\, \bm{\lambda}_x^{\mathrm{T}} \bm{M}\,\bm{\alpha}\, u.
  \end{aligned}
\end{equation}
where the detailed expressions of partial derivative of $\bm{M}$ and $\bm{D}$ to $\bm{x}$ are reported in supplementary material.

According to Pontryagin's Minimum Principle,
the Hamiltonian must be minimized with respect to the control, which leads to the
optimal thrust direction and magnitude:
\begin{equation}
  \label{eq:indirect_optimal_control}
  \bm{\alpha}^* = -\frac{\bm{M}^{\mathrm{T}} \bm{\lambda}_x}{\|\bm{M}^{\mathrm{T}} \bm{\lambda}_x\|},
  u^* = \frac{2\epsilon}{\rho+2\epsilon+\sqrt{\rho^2+4\epsilon^2}},
\end{equation}
where $\rho$ is the switching function and expressed by:

\begin{equation}
  \rho = 1 - \frac{I_{\rm sp} g_0 \|\bm{M}^{\mathrm{T}} \bm{\lambda}_x\| }{\lambda_0 m } - \frac{\lambda_m}{\lambda_0}.
\end{equation}

Since the terminal mass is free in the problem, the transversality condition is
provided by
  $\lambda_m(t_f) = 0.$
In addition, the
augmented costate vector
$\bm{\lambda}(t) \triangleq \big[\bm{\lambda}_x^\mathsf{T}(t),\, \lambda_m(t),\, \lambda_0\big]^\mathsf{T}$
is normalized at the initial time:
$  \|\bm{\lambda}(t_0)\| = 1.$
Consequently, the fuel-optimal control problem is converted into the following
two-point boundary value problem:
\begin{equation} 
  \label{eq:indirect_TPBVP}
  F_{\rm fuel}\left[\bm{\lambda}(t_0)\right] =
  \begin{bmatrix}
    \bm{x}(t_f) - \bm{x}_f \\
    \lambda_m(t_f)         \\
    \|\bm{\lambda}(t_0)\| - 1
  \end{bmatrix}
  = \bm{0}.
\end{equation}
This TPBVP is solved via a shooting method \cite{MoreJ.J.GarbowB.S.andHillstrom1980}. Once the optimal initial costate $\bm{\lambda}_x(t_0)$ and final time $t_f$ are determined, the state and costate equations \eqref{eq:indirect_dynamics} and \eqref{eq:indirect_costate} can be integrated to obtain the complete trajectory and corresponding control law. 

For the time optimal problem, only the differences are presented because of
majority similar derivations. The Hamiltonian is defined as
\begin{equation}
  H_{\rm time} =
  \frac{T_{\max}}{m} \bm{\lambda}_x^{\mathrm{T}} \bm{M}\,\bm{\alpha} u + \bm{\lambda}_x^{\mathrm{T}} \bm{D} - \lambda_m \frac{T_{\max}}{I_{\rm sp} g_0}u + \lambda_0.
\end{equation}
The costate dynamics and the optimal control direction remain the same as in
the fuel-optimal problem, expressed in
Eqs.(\ref{eq:indirect_costate}-\ref{eq:indirect_optimal_control}), while the
normalized magnitude of optimal control keeps the constant to 1 in time-optimal
problem.

With reference to the transversality conditions provided in
\cite{levineControlSystemsHandbook2018}, the optimal final time $t_f$ is
determined by
\begin{equation}
  H_{\rm time}(t_f) - \bm{\lambda}_x(t_f) \cdot \dot{\bm{x}}_f = H_{\rm time}(t_f) - \bm{\lambda}_L(t_f)\frac{\sqrt{\mu p_f}}{r_f^2} = 0.
\end{equation}
The term $\sqrt{\mu p_f}/r_f^2$ is the natural Keplerian angular rate of the target’s true longitude $L$ at the final state. $p_f$ is the final semilatus rectum, and $r_f$ is the final orbital radius. Since the specific angular momentum is $h_f=\sqrt{\mu p_f}$, one has $\dot L_f = h_f/r_f^2$ for unforced two-body motion.

As a result, the time-optimal control problem is converted into the following
two-point boundary value problem:
\begin{equation}
  \label{eq:indirect_TPBVP_time} 
  F_{\rm time}\left[\bm{\lambda}(t_0); t_f\right] =
  \begin{bmatrix}
    \bm{x}(t_f) - \bm{x}_f(t_f)                                            \\
    \lambda_m(t_f) = 0                                                \\
    H_{\rm time}(t_f) - \bm{\lambda}_L(t_f)\frac{\sqrt{\mu p_f}}{r_f^2} \\
    \|\bm{\lambda}(t_0)\| - 1
  \end{bmatrix}
  = \bm{0}.
\end{equation}

\subsection{Homotopy Ray Method for Large-Scale Data Generation}
\label{sec:homotopy_ray_method_for_large_scale_data_generation}
To tackle the challenge of generating large-scale datasets, the Homotopy Ray Method is proposed. 
Unlike existing homotopy methods, which continuously deform an easier auxiliary problem into a prescribed transfer, the present method uses the homotopy idea for dataset construction. Its objective is not only to solve one specific transfer problem, but also to distribute samples throughout the admissible state space while emphasizing critical regimes.
The primary goal is to efficiently generate samples that exhibit low fuel consumption or lie close to the reachable boundary.
This focus is essential because optimal solutions of numerical trajectory optimization typically lie in these extreme or boundary regions. If these critical regimes are underrepresented in the training data, a model may achieve low average test error while producing large estimation errors precisely in regions relevant to optimization. Such localized inaccuracies can severely degrade optimization performance, resulting in suboptimal or even infeasible trajectories. It is illustrated through a representative example in Sec.~\ref{sec:thirdparty}.

\subsubsection{Effects of Dataset Size and Model Parameter Scale.}

\begin{figure}[hbt!]
  \centering
  \includegraphics[width=0.58\textwidth]{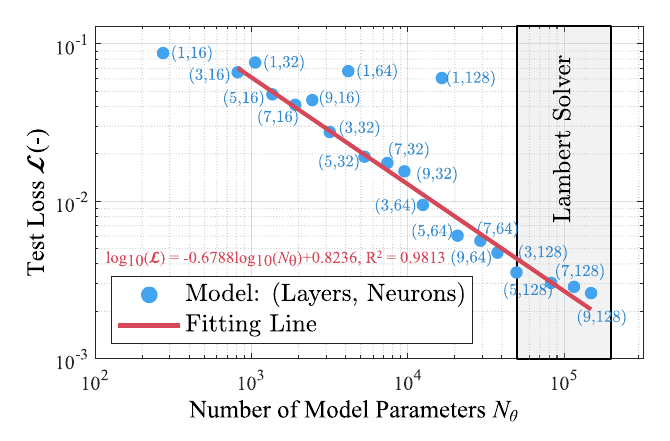}
  \caption{Model size scaling law in low-thrust approximation.}
  \label{fig:scalinglaw_model}
\end{figure}

\begin{figure}[hbt!]
  \centering
  \includegraphics[width=0.55\textwidth]{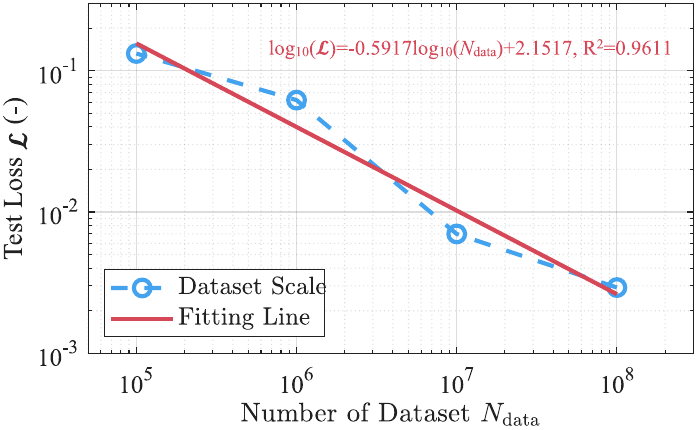}
  \caption{Dataset size scaling law in low-thrust approximation.}
  \label{fig:scalinglaw_datasize}
\end{figure}

The effect of dataset size and model capacity on approximation accuracy is first
examined using fuel-consumption prediction as a representative case. The neural
network inputs, outputs, and training procedure are described in
Sec.~\ref{sec:NNTraining}. Figures~\ref{fig:scalinglaw_model} and
\ref{fig:scalinglaw_datasize} summarize the resulting performance trends.
Figure~\ref{fig:scalinglaw_model} shows the dependence of the prediction error
on model size, where each point corresponds to a distinct network architecture
defined by its number of layers and neurons. As model capacity increases, the
loss decreases systematically, following an approximately linear trend on a
logarithmic scale. A similar behavior is observed in
Fig.~\ref{fig:scalinglaw_datasize}, which reports the effect of increasing the
training dataset size.  In both cases, no evidence of performance saturation is
observed within the explored ranges, indicating that approximation exhibits a clear scaling-law
behavior.

To make the scaling-law statement explicit, we fit the test loss
$\mathcal{L}$ with a log--log linear model,
\begin{equation}
\log_{10}(\mathcal{L}) = a - b \log_{10}(N),
\label{eq:scalinglaw_fit}
\end{equation}
where $N$ here denotes either the number of trainable parameters $N_\theta$ or the
number of training samples $N_{\mathrm{data}}$. 
The regressions shown in
Figs.~\ref{fig:scalinglaw_model} and \ref{fig:scalinglaw_datasize} exhibit strong
agreement with Eq.~\eqref{eq:scalinglaw_fit}, with coefficients of determination
exceeding 0.96 in both cases. These results indicate that
Eq.~\eqref{eq:scalinglaw_fit} captures most of the variance of the measured trends
within the tested ranges.

An important observation is that the effects of dataset size and model capacity
are largely independent: increasing the size of the training set does not alter
the optimal model scale, and vice versa. For each experimental setting, several
network configurations were evaluated (see Sec.~\ref{sec:NNTraining}), and the
best-performing hyperparameters were retained. The observed trends are
consistent in form with scaling behaviors reported in other large-scale neural
approximation problems, including large language models~\cite{kaplanScalingLawsNeural2020}.
We emphasize, however, that the fitted exponents above should be interpreted as
an empirical observation under the specific pipeline used in this work (input
features, normalization, and training protocol described in
Sec.~\ref{sec:NNTraining}), and whether a similar scaling law holds universally is beyond the scope
of the present study and warrants further investigation.

While increasing model size improves accuracy, it also increases inference
cost, which becomes critical when the network is embedded within
mission-level optimization frameworks. For this reason, model capacity is
constrained such that inference time remains on the same order of magnitude as
that of a Lambert solver~\cite{izzoRevisitingLambertsProblem2015}, as indicated
by the shaded region in Fig.~\ref{fig:scalinglaw_model}. This criterion leads to
the selection of a network with 9 hidden layers and 128 neurons per layer.
Beyond this point, further accuracy gains are pursued by increasing dataset
size, which affects training cost but does not impact runtime performance for users.

\subsubsection{Large-Scale Data Generation}
\label{sec:KeplerianNeighborhood}

\begin{figure}[hbt!]
  \centering
  \includegraphics[width=0.5\textwidth]{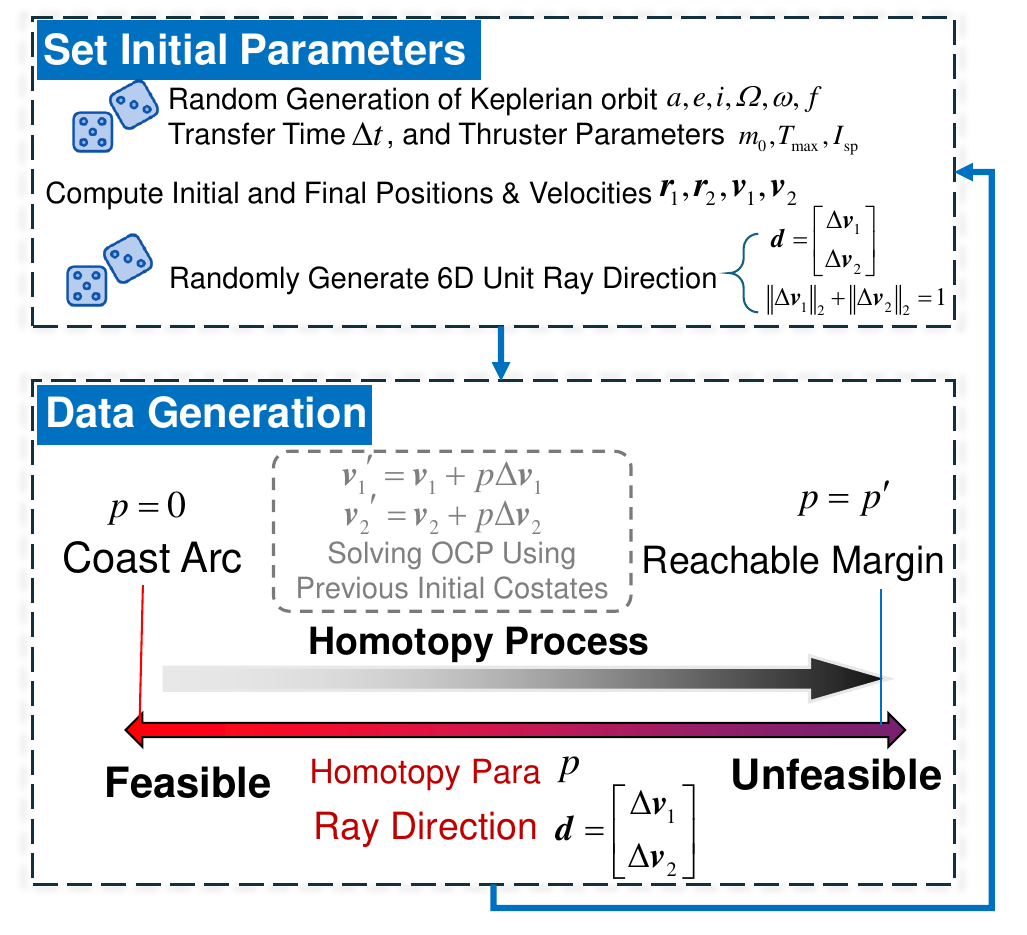}
  \caption{Schematic diagram of the homotopy ray method.}
  \label{fig:homotopy_generate}
\end{figure}

To further improve approximation accuracy, the
dataset is expanded to large scale using a structured data generation strategy.
The proposed approach exploits the fact that low-thrust transfers of practical
interest tend to cluster around feasible, near-optimal regions of the design
space. Rather than sampling boundary conditions uniformly, which results in a
high fraction of infeasible or uninformative cases, we generate trajectories by
progressively deforming guaranteed-feasible solutions toward the boundary of
reachability. This strategy, referred to as the \emph{Homotopy Ray Method}, is
illustrated schematically in Fig.~\ref{fig:homotopy_generate}.

The method consists of two steps. First, an initial feasible transfer is
constructed by sampling boundary conditions from a Keplerian orbital
neighborhood. Second, this solution is continuously deformed along a prescribed
direction in boundary-condition space until infeasibility is reached, while
reusing previously computed costates to ensure fast and reliable convergence of
the shooting method.

Initially, random initial guesses are generated 
from ballistic trajectories, which are trivially feasible even in the
absence of thrust. Under the two-body assumption, a Keplerian orbit is randomly
sampled using the orbital elements $(a,e,i,\Omega,\omega,f)$, a transfer
duration $\Delta t$ and other parameters $m_0, T_{\max}, I_{\rm sp}$,  and converted into the corresponding boundary states
$(\boldsymbol{r}_1,\boldsymbol{v}_1,\boldsymbol{r}_2,\boldsymbol{v}_2)$. These
states serve as reliable starting points for subsequent homotopic deformation.

Then, starting from a Keplerian solution, a homotopy ray is defined in the space of
boundary velocities by a random, normalized six-dimensional direction vector
$d = [\Delta\boldsymbol{v}_1;\Delta\boldsymbol{v}_2]$, with
$\|\Delta\boldsymbol{v}_1\|^2 + \|\Delta\boldsymbol{v}_2\|^2 = 1$. A homotopy
parameter $p$ controls the deviation from the original Keplerian boundary
conditions, such that
$
\boldsymbol{v}_1' = \boldsymbol{v}_1 + p\,\Delta\boldsymbol{v}_1,
\boldsymbol{v}_2' = \boldsymbol{v}_2 + p\,\Delta\boldsymbol{v}_2 .
$
Beginning from a small initial value of $p$, the parameter is gradually
increased, generating a sequence of trajectory optimization problems along the
homotopy ray. 
The continuation proceeds until the trajectory becomes infeasible or numerical
convergence can no longer be achieved.

Repeating this procedure for multiple Keplerian orbits and homotopy directions
yields a large set of high-quality trajectories that densely populate the
feasible region and its boundary. Algorithm~\ref{alg:HomotopyRay} summarizes the
implementation details of the Homotopy Ray Method.

\begin{algorithm}[htbp!]
  \caption{Homotopy Ray Method}
  \label{alg:HomotopyRay}
  \begin{algorithmic}[1]
    \REPEAT
    \STATE Randomly generate initial position $\boldsymbol{r}_1$, velocity $\boldsymbol{v}_1$, terminal position $\boldsymbol{r}_2$, velocity $\boldsymbol{v}_2$, transfer time $\Delta t$, initial mass $m$, specific impulse $I_{\rm sp}$ and thrust $T_{\max}$
    \STATE Randomly generate normalized perturbations $\Delta \boldsymbol{v}_1$ and $\Delta \boldsymbol{v}_2$ such that $\|\Delta \boldsymbol{v}_1\|^2 + \|\Delta \boldsymbol{v}_2\|^2 = 1$
    \STATE Initialize the homotopy parameter $p_{\rm start}$ and solve the initial costate $\boldsymbol{\lambda}_0$ for $p_{\rm start}$
    \STATE Initialize the empty set of solved trajectories $\mathcal{S}_{\rm solved}$
    \REPEAT
    \STATE Compute perturbed velocities:
    $
      \boldsymbol{v}_1' = \boldsymbol{v}_1 + p\,\Delta \boldsymbol{v}_1, \quad \boldsymbol{v}_2' = \boldsymbol{v}_2 + p\,\Delta \boldsymbol{v}_2
    $
    \STATE Solve the TPBVP using the current costate guess $\boldsymbol{\lambda}_0$
    \IF{the TPBVP is solved successfully}
    \STATE Update: $p \gets p + \Delta p$
    \STATE Set $\boldsymbol{\lambda}_0$ to the newly obtained costate for the next iteration
    \STATE Store the trajectory in $\mathcal{S}_{\rm solved}$
    \ELSE
    \STATE Reduce the step size: $\Delta p \gets \Delta p/2$
    \ENDIF
    \UNTIL{the problem becomes unsolvable: $\Delta p$ falls below a pre-defined threshold}
    \STATE Select trajectories from $\mathcal{S}_{\rm solved}$ to move to the Data Sets $\mathcal{S}$
    \UNTIL{sufficient data is generated}
    \STATE Output Data Sets $\mathcal{S}$
  \end{algorithmic}
\end{algorithm}

The physical meaning of infeasibility differs between the two optimal control
formulations considered in this work. For the fuel-optimal problem, the maximum
value of the homotopy parameter corresponds to the boundary of the reachable
set; in this case, the computed time-optimal transfer duration coincides with
$\Delta t$. For the time-optimal problem, infeasibility typically arises from
excessive fuel consumption, as the engine operates at maximum thrust throughout
the transfer. This may lead to highly eccentric or even hyperbolic trajectories,
causing numerical difficulties. To avoid such singular cases, an early
termination condition is introduced for the time-optimal formulation. Specifically, the optimization process is terminated when either the initial or final orbit becomes hyperbolic, or when the remaining mass fraction $m_f/m_0$ falls below 0.4.

\subsubsection{Data Distribution Characteristics During Homotopy}

\begin{figure}[hbt!]
  \centering
  \includegraphics[width=0.75\linewidth]{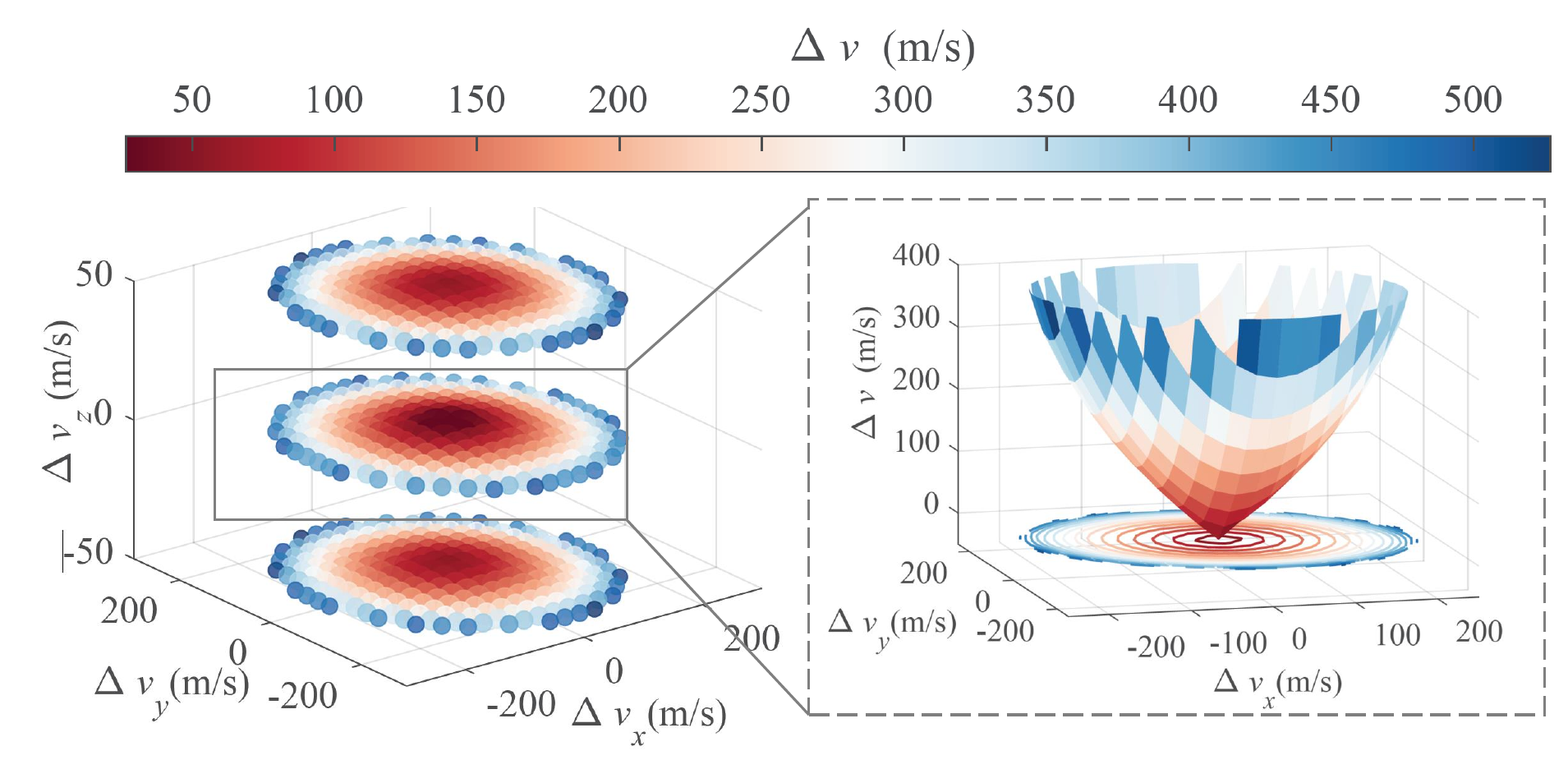}
  \caption{Relationship between fuel consumption and the corresponding variation in departure velocity.}
  \label{fig:Figure_convex_diagram}
\end{figure}

The homotopy ray method induces a structured data distribution with an approximately monotonic relationship between the velocity perturbations $(\Delta\boldsymbol{v}_1,\Delta\boldsymbol{v}_2)$ and the resulting fuel consumption $\Delta v$, as shown in Fig.~\ref{fig:Figure_convex_diagram}. Starting from a feasible Keplerian solution, larger boundary perturbations move the trajectory toward the reachability boundary and require progressively higher control effort. Consequently, samples along each homotopy ray vary smoothly and are more ordered than those obtained from random sampling.
This structure motivates the use of velocity increments $(\Delta\boldsymbol{v}_1,\Delta\boldsymbol{v}_2)$ as network inputs, since they encode the transfer difficulty relative to a feasible reference. Similar conclusions have been reported
in~\cite{acciariniComputingLowthrustTransfers2024a,liDeepNetworksApproximators2020,zhuFastEvaluationLowThrust2019},
where Lambert-based solutions were employed as input features to improve network
performance. By leveraging the homotopy parameter $p$, the method continuously transitions solutions from a feasible reference transfer toward the reachability boundary. This process inherently concentrates sampling in the low-fuel regime and near the boundary of reachability, which are of primary interest in mission design.

\section{Data Preparation}
\label{sec:input_output_analysis}

The objective of this section is to
identify a minimal set of independent variables that preserves the full physical
content of the original problem.
To reduce the input dimensionality of model and improve learning efficiency, first, dimensionality reduction is
introduced in a self-similar space by exploiting the inherent invariance
properties of the low-thrust transfer problem. Next, the influence of different
input configurations on model performance is evaluated. Finally, the complete
input feature sets for the two neural network models, along with intermediate
data processing steps, are summarized.

The original input 
variables defining the optimal control problems include
$
\{\boldsymbol{r}_1,\boldsymbol{v}_1,\boldsymbol{r}_2,\boldsymbol{v}_2,\Delta t,
m_0,T_{\max},I_{\mathrm{sp}},\mu\}.
$
Here, Cartesian coordinates are used for clarity, although  the same analysis applies
when modified equinoctial elements are employed. In total, the problem
contains 17 independent input variables. $g_0$ is not treated as an independent parameter because it always
appears in the product $g_0 I_{\mathrm{sp}}$. Any variation in $g_0$ can
therefore be absorbed into an equivalent change in $I_{\mathrm{sp}}$, and
$g_0$ is fixed at the standard value of 9.80665~m/s$^2$ throughout this work.

\subsection{Dimensionality Reduction in Self-Similar Space}
\label{sec:dimensionality_reduction}

Dimensionality reduction is achieved by exploiting the rotational and
dimensional invariance. It should be
emphasized that these transformations correspond to data preprocessing rather
than data augmentation for neural network training. Since they preserve the physical equivalence of the
underlying transfer, no information relevant to the optimal solution is lost.

\subsubsection{Rotational Invariance}

Under two-body dynamics, the equations of motion depend only on the
relative position with respect to the central body and are invariant under
global coordinate rotation. If the initial and terminal states
$(\boldsymbol{r}_1,\boldsymbol{v}_1,\boldsymbol{r}_2,\boldsymbol{v}_2)$ are
simultaneously rotated by the same transformation, the dynamics, constraints,
and optimality conditions remain unchanged.
As a result, the optimal solution, whether minimizing fuel consumption or
transfer time, is invariant under coordinate rotation. This property guarantees
that the physical characteristics of the transfer are independent of the chosen
reference frame and allows the problem to be expressed in a reduced coordinate
system without loss of generality.

\begin{figure}[htb!]
  \centering
  \includegraphics[width=0.4\linewidth]{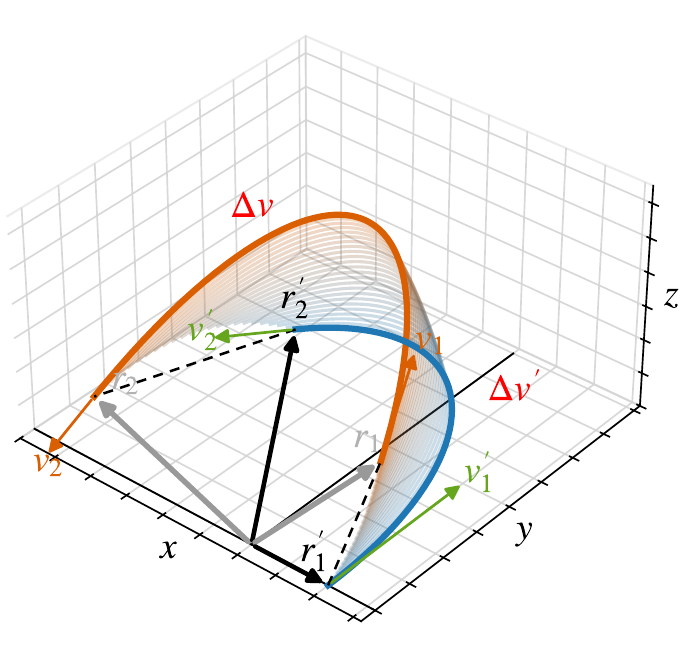}
  \caption{The diagram illustrates the rotational invariance.}
  \label{fig:dimensionless_rotational}
\end{figure}

As illustrated in Fig.~\ref{fig:dimensionless_rotational}, rotating the boundary
states yields transformed variables
$(\boldsymbol{r}_1',\boldsymbol{v}_1',\boldsymbol{r}_2',\boldsymbol{v}_2')$.
Because the governing equations are rotationally invariant, the resulting
optimal velocity increment satisfies $\Delta v' = \Delta v$, with an analogous
result holding for time-optimal transfers.
In this work, the rotational degrees of freedom are eliminated by aligning the
departure position with the $x$-axis and constraining the arrival position to
lie in the $xy$-plane. This transformation removes three redundant degrees of
freedom while preserving the full physical structure of the transfer.

The only remaining ambiguity is the choice between the two mirror orientations
of the reduced frame. We fix this convention by selecting the orientation for
which the rotated departure velocity has a positive transverse component,
$
(\boldsymbol v_1')_y>0,
$
thereby assigning a unique representative to each boundary condition.

\subsubsection{Dimensional Invariance}

Dimensional invariance is widely exploited in low-thrust trajectory
optimization to improve numerical conditioning and expose fundamental scaling
properties. Typical approaches include adopting canonical units such as
1~AU and 1~year~\cite{jiangPracticalTechniquesLowThrust2012a}, or rescaling the
dynamics such that the gravitational parameter becomes unity~\cite{izzoRealTimeGuidanceLowThrust2021}.
In this work, dimensional invariance is employed for input parameterization,
with emphasis on selecting appropriate dimensional quantities as independent
inputs to the neural network.

\begin{figure}[htbp]
  \centering
  \includegraphics[width=0.4\linewidth]{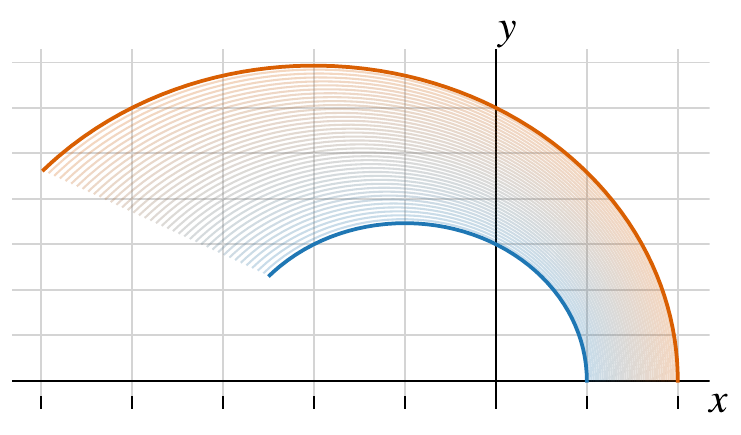}
  \caption{Illustration of dimensional invariance.}
  \label{fig:dimensionless}
\end{figure}

The dimensional normalization is applied after the rotational reduction
introduced above. Let
$(\boldsymbol r_1',\boldsymbol v_1',
\boldsymbol r_2',\boldsymbol v_2')$
denote the rotated boundary states. As illustrated in
Fig.~\ref{fig:dimensionless}, with
$L_0 \triangleq \|\boldsymbol r_1'\|=\|\boldsymbol r_1\|$,
the transformed position vectors take the form
$\boldsymbol r_1'=[L_0,0,0]^{\mathsf T}$ and
$\boldsymbol r_2'=[x_2',y_2',0]^{\mathsf T}$.
Thus, the rotational reduction fixes the coordinate orientation, while the
remaining dimensional scale is represented by $L_0$. The corresponding
Keplerian time scale is chosen as
$T_0 \triangleq \sqrt{L_0^3/\mu}$, with induced velocity and acceleration units
$V_0=L_0/T_0$ and $A_0=L_0/T_0^2$. 

The boundary variables after both reductions are therefore
$\tilde{\boldsymbol r}_1=[1,\,0,\,0]^{\mathsf T}$,
$\tilde{\boldsymbol r}_2=\frac{1}{L_0}[x_2',\,y_2',\,0]^{\mathsf T}$,
$\tilde{\boldsymbol v}_i=\boldsymbol v_i'/V_0$ $(i=1,2)$,
and $\Delta\tilde t=\Delta t/T_0$.
This construction makes explicit that the coordinate orientation, absolute
length scale, and central-body gravitational parameter are not independent
input features; instead, they define the representative normalized
coordinates.

Using the characteristic scales defined above, the trajectory variables are
nondimensionalized as
\begin{equation}
  \label{eq:nd_vars}
  \tilde t=\frac{t}{T_0},\qquad
  \tilde{\bm r}=\frac{\bm r}{L_0},\qquad
  \tilde{\bm v}=\frac{\bm v}{V_0}
  =\frac{T_0}{L_0}\bm v,\qquad
  \tilde m=\frac{m}{m_0}.
\end{equation}
To parameterize the propulsion system, define
$a_s \triangleq T_{\max}/m_0$
and
$c \triangleq I_{\mathrm{sp}}g_0$. Their nondimensional counterparts are
$\bar{a}_s \triangleq a_s/A_0$
and
$\bar{c} \triangleq c/V_0=I_{\mathrm{sp}}g_0/V_0$.
The resulting normalized equations of motion are
\begin{equation}
  \label{eq:nd_dynamics}
  \frac{d\tilde{\bm r}}{d\tilde t}
  =\tilde{\bm v},\qquad
  \frac{d\tilde{\bm v}}{d\tilde t}
  =-\frac{\tilde{\bm r}}{\tilde r^3}
  +\bar{a}_s
  \frac{u\boldsymbol\alpha}{\tilde m},\qquad
  \frac{d\tilde m}{d\tilde t}
  =-\frac{\bar{a}_s}{\bar{c}}u.
\end{equation}

Equation~\eqref{eq:nd_dynamics} shows that, after normalization, the transfer
dynamics no longer depend explicitly on the absolute scale $L_0$ or the
gravitational parameter $\mu$. Instead, the propulsion system is characterized
by the normalized acceleration $\bar a_s$ and exhaust velocity $\bar c$.
Consequently, the dimensional variables
$(L_0,\mu,T_{\max},m_0,I_{\mathrm{sp}})$ are replaced by the compact feature set
$(\bar a_s,\bar c)$ together with the normalized boundary conditions.

In summary, the original formulation involves 17 independent variables. By
accounting for rotational invariance (three degrees of freedom) and dimensional
invariance (three degrees of freedom: $L_0, \mu, a_s$), the effective input dimensionality is
reduced to 11 without loss of physical information.

\subsection{Comparison of Different Types of Inputs}
\label{sec:comparison_different_inputs}

Building on the minimum set of independent variables, this subsection evaluates the effect of different input representations on neural network performance. 
The considered configurations, which are commonly used in low-thrust trajectory optimization and learning, are summarized in Table~\ref{tab:input_params}.

\begin{table}[htbp]
  \centering
  \caption{Input Parameters and Descriptions}
  \begin{tabular}{ll}
    \toprule
    Input                                                 & Description                                                                                                               \\
    \midrule
    $\mathrm{coe}$                                        & $\text{coe}_1,\ \text{coe}_2$                                                                                             \\
    $\mathrm{mee}$                                        & $\text{mee}_1,\ \text{mee}_2$                                                                                             \\
    $\mathrm{rv}$                                         & $\boldsymbol{r}_1,\ \boldsymbol{v}_1,\ \boldsymbol{r}_2,\ \boldsymbol{v}_2$                                               \\
    $\mathrm{rv_{rotate}}$                                & $\boldsymbol{r}_1',\ \boldsymbol{v}_1',\ \boldsymbol{r}_2',\ \boldsymbol{v}_2'$                                           \\
    Ref.~\cite{acciariniComputingLowthrustTransfers2024a} & $\boldsymbol{r}_1-\boldsymbol{r}_2,\ \boldsymbol{v}_1-\boldsymbol{v}_2,\ \Delta\boldsymbol{v}_1,\ \Delta\boldsymbol{v}_2$ \\
    $\mathrm{Lambert_{cart}}$                             & $e,\ f,\ \Delta\boldsymbol{v}_1,\ \Delta\boldsymbol{v}_2$ (in $[x,y,z]^\mathrm{T}$)                                       \\
    $\mathrm{Lambert_{sph}}$                              & $e,\ f,\ \Delta\boldsymbol{v}_1,\ \Delta\boldsymbol{v}_2$ (in $[r,\theta,\varphi]^\mathrm{T}$)                            \\
    $t_\mathrm{Lambert}$                                  & $(\|\Delta\boldsymbol{v}_1\|+\|\Delta\boldsymbol{v}_2\|)/a_s$                                                             \\
    \bottomrule
  \end{tabular}
  \label{tab:input_params}
\end{table}

In Table~\ref{tab:input_params}, $\Delta t$, $a_s$, and $I_{\mathrm{sp}}$ are included in every configuration and are therefore omitted. All variables are normalized, and quantities that become constant after the invariance transformations are excluded; for example, $\boldsymbol{r}_1'$ is fixed to $[1,0,0]^{\mathrm{T}}$ in the $\mathrm{rv_{rotate}}$ case. Angular variables are represented by their sine and cosine to preserve periodicity, and Lambert-based inputs are obtained from the corresponding Lambert problem.
\begin{table}[htbp]
  \centering
  \caption{$\Delta v$ Prediction Performance Comparison of Different Input Configurations}
  \begin{tabular}{cccccc}
    \toprule
                                                          & Train  & Val    & Test   & $e_{\rm abs}$, & $e_{\rm rel}$, \\
    Input                                                 & Loss   & Loss   & Loss   & m/s            & (\%)           \\
    \midrule
    $\mathrm{coe}$                                        & 0.0012 & 0.0034 & 0.0032 & 3.76           & 6.21           \\
    $\mathrm{mee}$                                        & 0.0010 & 0.0030 & 0.0034 & 3.94           & 6.49           \\
    $\mathrm{rv}$                                         & 0.0096 & 0.0120 & 0.0113 & 13.21          & 27.52          \\
    $\mathrm{rv_{rotate}}$                                & 0.0005 & 0.0026 & 0.0024 & 2.83           & 4.71           \\
    Ref.~\cite{acciariniComputingLowthrustTransfers2024a} & 0.0001 & 0.0021 & 0.0019 & 2.26           & 1.30           \\
    $\mathrm{Lambert_{cart}}$                             & 0.0002 & 0.0019 & 0.0017 & 2.00           & 3.58           \\
    $\mathrm{Lambert_{sph}}$                              & 0.0002 & 0.0020 & 0.0018 & 2.11           & 1.07           \\
    $\mathrm{rv_{rotate}},\mathrm{Lambert_{sph}}$         & 0.0002 & 0.0021 & 0.0019 & 2.19           & 1.18           \\
    \bottomrule
  \end{tabular}%
  \label{tab:comparison_inputs_dv}%
\end{table}%

The prediction performance of different input configurations is evaluated for
two separate networks: one for fuel consumption ($\Delta v$) prediction and the
other for transfer time ($\Delta t$) prediction. The results are summarized in
Tables~\ref{tab:comparison_inputs_dv}
and~\ref{tab:comparison_inputs_dt}.
To ensure a controlled comparison, all experiments use the same network
architecture consisting of nine hidden layers with 128 neurons per layer. The
training dataset contains 200{,}000 samples, and only single-revolution
transfers are considered. For each input configuration, hyperparameters are
independently tuned, as described in Sect.~\ref{sec:NNTraining}.

\begin{table}[htbp]
  \centering
  \caption{$\Delta t$ Prediction Performance Comparison of Different Input Configurations}
  \begin{tabular}{c
      S[table-format=1.4, round-mode=places, round-precision=4] 
      S[table-format=1.4, round-mode=places, round-precision=4] 
      S[table-format=1.4, round-mode=places, round-precision=4] 
      S[table-format=1.2, round-mode=places, round-precision=2] 
      S[table-format=1.2, round-mode=places, round-precision=2] 
    }
    \toprule
                                                                     & {Train} & {Val}   & {Test}  & {$e_{\rm abs}$,} & {$e_{\rm rel}$,} \\
    {Input}                                                          & {Loss}  & {Loss}  & {Loss}  & {days}           & {(\%)}           \\
    \midrule
    $\mathrm{coe}$                                                   & 0.00119 & 0.00722 & 0.00691 & 3.1069           & 1.516            \\
    $\mathrm{mee}$                                                   & 0.00030 & 0.00606 & 0.00501 & 2.2540           & 1.059            \\
    $\mathrm{rv}$                                                    & 0.02512 & 0.03228 & 0.01680 & 7.5541           & 6.071            \\
    $\mathrm{rv_{rotate}}$                                           & 0.00015 & 0.00648 & 0.00452 & 2.0321           & 0.843            \\
    Ref.~\cite{acciariniComputingLowthrustTransfers2024a}            & 0.00016 & 0.00607 & 0.00451 & 2.0280           & 0.723            \\
    $\mathrm{Lambert_{cart}}$                                        & 0.00015 & 0.00537 & 0.00392 & 1.7619           & 0.742            \\
    $\mathrm{Lambert_{sph}}$                                         & 0.00017 & 0.00494 & 0.00399 & 1.7918           & 0.726            \\
    $\mathrm{rv_{rotate}},t_\mathrm{Lambert}$                        & 0.00011 & 0.00426 & 0.00415 & 1.8670           & 0.719            \\
    $\mathrm{Lambert_{sph}},t_\mathrm{Lambert}$                      & 0.00013 & 0.00650 & 0.00409 & 1.8370           & 0.699            \\
    $\mathrm{rv_{rotate}},\mathrm{Lambert_{sph}},t_\mathrm{Lambert}$ & 0.00009 & 0.00424 & 0.00402 & 1.8097           & 0.685            \\
    \bottomrule
  \end{tabular}%
  \label{tab:comparison_inputs_dt}%
\end{table}%

Several conclusions can be drawn from Tables~\ref{tab:comparison_inputs_dv} and~\ref{tab:comparison_inputs_dt}. First of all, when Lambert information is not used, $\mathrm{rv_{rotate}}$ consistently outperforms the other state representations, while MEE performs best among the orbital-element inputs and raw Cartesian position-velocity performs worst. Adding Lambert-derived features improves both $\Delta v$ and $\Delta t$ prediction, showing that these quantities provide strong physical priors. In addition, Lambert-based inputs alone already achieve competitive performance, and appending $\mathrm{rv_{rotate}}$ does not consistently improve the results. Across all tested cases, spherical Lambert features outperform Cartesian ones.

For the time-optimal network, transfer time is included as an input solely because the model is used to assess reachability for a prescribed time of flight. The formulation used here is therefore a reachability-oriented one, and the full time-optimal case without transfer time as an input is deferred to Sec.~\ref{sec:NNTraining_MultiRev}.

\section{Neural Network Training: Single-Revolution Transfers}
\label{sec:NNTraining}

This section investigates the impact of training-related parameters on the predictive performance of deep neural networks for single-revolution transfers. The discussion is divided into two parts. First, we evaluate multiple network architectures under a range of hyperparameter settings, reporting only the best result for each configuration. Second, we analyze the sensitivity of the optimal architecture to individual hyperparameters.

\begin{figure}[hbt!]
  \centering
  \includegraphics[width=0.7\linewidth]{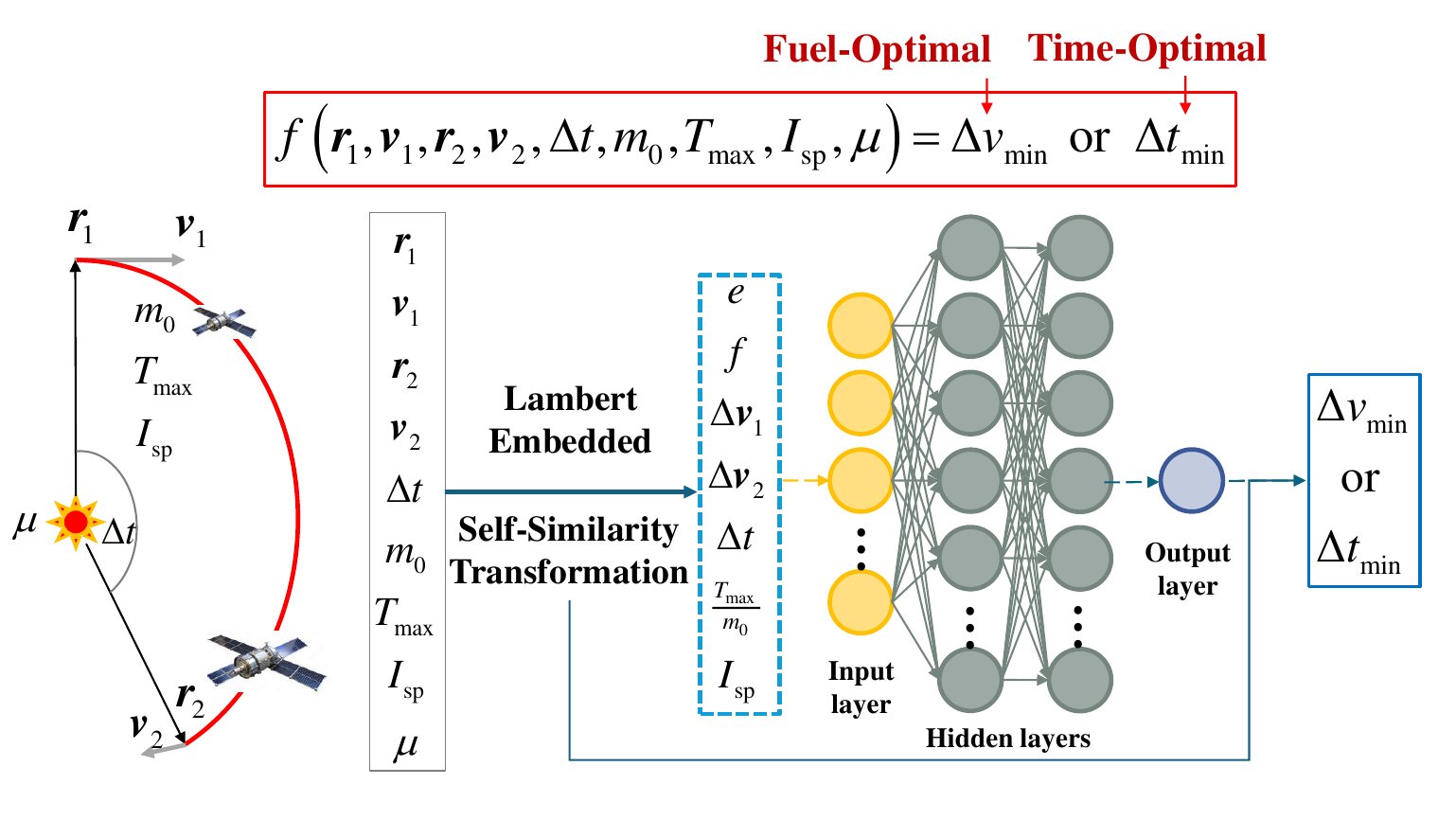}
  \caption{The input and output of the prediction model.}
  \label{fig:input_output}
\end{figure}

The dataset is split into training, validation, and test sets using a 96\%/2\%/2\% partition. Each model is trained for 10,000 epochs, with validation performance monitored at every epoch. The model achieving the best validation performance is retained to mitigate overfitting and subsequently evaluated on the test set to assess generalization. To efficiently explore a large hyperparameter space, all experiments are conducted on a subset of 100,000 samples.
Following the findings in Sec.~\ref{sec:comparison_different_inputs}, the $\mathrm{Lambert_{sph}}$ input configuration is selected. While including $t_{\mathrm{Lambert}}$ improves performance for the time-optimal model, $\mathrm{Lambert_{sph}}$ ensures consistent input design across both $\Delta v$ and $\Delta t$ models.

The overall input-output pipeline is illustrated in Fig.~\ref{fig:input_output}. Raw mission parameters are first processed to enforce rotational invariance and non-dimensionalization, reducing the effective input dimensionality. A Lambert solver then computes the corresponding two-impulse transfer characteristics, which serve as inputs to the neural network. The network predicts dimensionless outputs, which are rescaled using the original velocity and time units to yield the final $\Delta v$ and minimum transfer duration.
All training is performed on a single NVIDIA RTX 4090 GPU (24 GB) with access to 6 CPU cores and 60 GB RAM. The models are implemented in Python 3.12 with PyTorch 2.5.1. Data generation is executed in C++ on a workstation with an AMD EPYC 7452 processor (2.6 GHz, 64 cores) and 256 GB RAM.

\subsection{Neural Network Structure}

Network architecture strongly influences predictive performance. Here, the number of hidden layers ($n_{\rm layer}$) and neurons per layer ($n_{\rm neuron}$) are treated as tunable hyperparameters.
 Each hidden layer uses a ReLU activation, and the output layer is linear. 

\begin{table}[htb!]
  \centering
  \caption{ $\Delta v$ Performance for Different Layer and Neuron Configurations}
  \label{tab:nn_structure_dv}
  \adjustbox{max width=\textwidth}{
    \begin{tabular}{cccccccccccccccc}
      \toprule
                      & \multicolumn{3}{c}{Neurons = 8} & \multicolumn{3}{c}{Neurons = 16} & \multicolumn{3}{c}{Neurons = 32} & \multicolumn{3}{c}{Neurons = 64} & \multicolumn{3}{c}{Neurons = 128}                                                                                         \\
      \cmidrule{2-16} & Train                           & Test                             & Time,                            & Train                            & Test                              & Time, & Train  & Test   & Time, & Train   & Test    & Time, & Train  & Test   & Time, \\
      Layers          & {Loss}                          & {Loss}                           & {s}                              & {Loss}                           & {Loss}                            & {s}   & {Loss} & {Loss} & {s}   & {Loss}  & {Loss}  & {s}   & {Loss} & {Loss} & {s}   \\
      \midrule
      1               & 0.090                           & 0.092                            & 156                              & 0.085                            & 0.087                             & 154   & 0.073  & 0.076  & 155   & 0.063   & 0.067   & 154   & 0.057  & 0.061  & 155   \\
      3               & 0.073                           & 0.072                            & 223                              & 0.058                            & 0.066                             & 223   & 0.026  & 0.028  & 225   & {0.006} & {0.009} & {225} & 0.001  & 0.004  & 225   \\
      5               & 0.070                           & 0.074                            & 291                              & 0.044                            & 0.048                             & 292   & 0.017  & 0.019  & 294   & 0.004   & 0.006   & 293   & 0.001  & 0.003  & 295   \\
      7               & 0.070                           & 0.073                            & 360                              & 0.038                            & 0.041                             & 363   & 0.013  & 0.018  & 363   & 0.003   & 0.006   & 362   & 0.001  & 0.003  & 363   \\
      9               & 0.070                           & 0.070                            & 425                              & 0.040                            & 0.044                             & 426   & 0.012  & 0.016  & 431   & 0.002   & 0.005   & 430   & 0.001  & 0.003  & 431   \\
      \bottomrule
    \end{tabular}%
  }
\end{table}

Table~\ref{tab:nn_structure_dv} summarizes the performance of the $\Delta v$ model across different layer and neuron configurations. Each architecture is fine-tuned over other hyperparameters, and only the best result is reported. The performance consistently improves with increasing network depth and width, confirming the expected scaling behavior. Similar observations are obtained for the $\Delta t$ model; however, due to space limitations, the corresponding results are not repeated here.

\subsection{Hyperparameter Tuning}
\label{sec:hyperparameter_tuning}

Hyperparameter tuning is performed by jointly optimizing three key parameters: the learning rate ($\eta$), the weight decay ratio ($\rho = w_d/\eta$), and the batch size ($B$) for both $\Delta v$ and $\Delta t$ models.
The AdamW optimizer~\cite{loshchilovDecoupledWeightDecay2018,llugsiComparisonAdamAdaMax2021} is adopted for its robustness in training, as it decouples weight decay from the gradient update. 
A OneCycleLR scheduler~\cite{smithSuperconvergenceVeryFast2019} is employed to dynamically adjust the learning rate throughout training.
All hyperparameter experiments use the same network architecture: 9 hidden layers with 128 neurons each. Table~\ref{tab:hyperparameter} reports the training and test losses and runtime for both $\Delta v$ and $\Delta t$. For clarity, it lists ablation results obtained by varying one hyperparameter while fixing the others at their task-specific selected values. The selected best settings are shown in the last row.

\begin{table}[htb!]
  \centering
  \begin{threeparttable}
  \caption{Hyperparameter tuning for $\Delta v$ and $\Delta t$ models}
  \label{tab:hyperparameter}
  \begin{tabular}{@{}llcccccc@{}}
    \toprule
          &        &
          \multicolumn{3}{c}{$\Delta v$} &
          \multicolumn{3}{c}{$\Delta t$} \\
    \cmidrule(lr){3-5}\cmidrule(l){6-8}
    Varied\tnote{a}
          & Value(s)
          & \shortstack{Train\\loss}
          & \shortstack{Test\\loss}
          & \shortstack{Time\\(s)}
          & \shortstack{Train\\loss}
          & \shortstack{Test\\loss}
          & \shortstack{Time\\(s)} \\
    \midrule
    $\eta$
          & 0.1
          & 0.0219 & 0.0226 & 424
          & 0.0194 & 0.0212 & 426 \\
          & 0.001
          & 0.0140 & 0.0160 & 424
          & 0.0092 & 0.0114 & 425 \\
          & 0.0001
          & 0.0223 & 0.0247 & 421
          & 0.0153 & 0.0183 & 427 \\
    \midrule
    $\rho$
          & $\rho_{\Delta v}=0.1,\ \rho_{\Delta t}=0.01$
          & 0.0060 & 0.0082 & 424
          & 0.0033 & 0.0045 & 426 \\
    \midrule
    $B$
          & 6400
          & 0.0067 & 0.0092 & 221
          & 0.0035 & 0.0045 & 223 \\
          & 12800
          & 0.0081 & 0.0103 & 115
          & 0.0043 & 0.0053 & 115 \\
          & 25600
          & 0.0090 & 0.0125 & 62
          & 0.0051 & 0.0068 & 62 \\
          & 51200
          & 0.0103 & 0.0127 & 37
          & 0.0063 & 0.0080 & 37 \\
    \midrule
    Selected
          & \begin{tabular}[c]{@{}l@{}}
            $B={3200},\eta={0.01}$\\
              $\Delta v$:  $\rho_{\Delta v}={0.01}$ \\
              $\Delta t$:  $\rho_{\Delta t}={0.1}$ 
            \end{tabular}
          & 0.0062 & {0.0079} & 424
          & 0.0033 & {0.0044} & 426 \\
    \bottomrule
  \end{tabular}
  \begin{tablenotes}
    \footnotesize
    \item[a] Unlisted hyperparameters are fixed to the corresponding selected values of the same task shown in the last row.
  \end{tablenotes}
  \end{threeparttable}
\end{table}

The results indicate that the learning rate $\eta$ has the largest effect on model performance, while the weight decay ratio $\rho$ has a comparatively minor impact. Due to the low input dimensionality and moderate network size, full-batch processing is efficiently parallelized on the GPU. Consequently, training time scales inversely with batch size $B$, and smaller batches generally improve generalization. This introduces a trade-off between computational efficiency and model performance for large datasets. For this study, $B=6400$ is selected to balance training efficiency and scalability to datasets with up to 100 million samples.
Finally, while scaling laws (Sec.~\ref{sec:DataGeneration}) guide network design, these results highlight the continued importance of careful hyperparameter selection. Inappropriate choices, such as an excessively high learning rate, can cause orders-of-magnitude degradation in predictive performance.

\section{Neural Network Training: Multi-Revolution Transfers}
\label{sec:NNTraining_MultiRev}

Extending the work of Acciarini et al.~\cite{acciariniComputingLowthrustTransfers2024a}, which demonstrated that neural surrogates could be used in transfers with multiple revolutions, we generalize our neural network approximators to handle multi-revolution trajectories. This section describes the methodology for data generation, network training, and performance evaluation in the multi-revolution context. As most aspects are unchanged from the single-revolution case, we focus on the differences.

\subsection{Multi-Revolution Optimal Control Problem}

Data generation follows the methodology outlined in Sec.~\ref{sec:DataGeneration}, using the same homotopy continuation procedure. A defining feature of multi-revolution transfers is the existence of multiple local optima due to multiple revolutions,  which can make the input--output mapping ambiguous if different branches coexist for the same boundary conditions. In the fuel-optimal setting, Lambert-based features are available (given a prescribed transfer time), which allows the revolution count to be specified a priori. In contrast, in the multi-revolution minimum-time setting the final time is unknown and the revolution count is an outcome of the optimization, so no such prior information can be imposed.

\subsubsection{Fuel-Optimal Transfers}

For fuel-optimal transfers, the procedure is executed independently for each target number of revolutions, ensuring balanced representation across all cases. Although the existence of multiple distinct solutions even in the single revolution setting remains an open question, substantial numerical evidence suggests that, for a fixed revolution, the fuel optimal solution is typically unique; this assumption is also adopted in prior studies~\cite{taheri_how_2020,zhangGTOC11Results2023a}. Under this assumption, our homotopy continuation follows a consistent solution branch along each ray and therefore produces single valued training labels. This makes it well defined to train one network for each revolution count. 

The resulting dataset comprises millions of trajectories spanning zero to three revolutions (roughly 3 million per revolution count), partitioned into 80\% training, 10\% validation, and 10\% test samples.

The dataset covers a broad range of mission scenarios: eccentricities from 0 to 1, inclinations from $0^\circ \text{ to } 180^\circ$, semi-major axes spanning below 1 to above 5 AU, specific impulses from 2,000 s to over 5,000 s, and fuel costs ranging from nearly ballistic ($m_f/m_0\approx1$) to highly demanding transfers.
 The network inputs remain those identified as optimal in Sec.~\ref{sec:input_output_analysis}: $\mathrm{rv_{rotate}}$ and $\mathrm{Lambert_{sph}}$.

\subsubsection{Time-Optimal Transfers}

In multi-revolution time-optimal problems, the transfer time $t_f$ is an optimization variable and is therefore unknown a priori. 
This introduces two challenges for data generation and supervised learning. First, along a continuation path, the shooting method may transition between different revolution optimal solutions. Second, multiple locally optimal transfer times may correspond to the same input, leading to label ambiguity and complicating network training.

To address these challenges, we adopt a two-step procedure. 
First, the original point-to-point terminal constraints are reformulated as orbital transfer terminal constraints, where only the terminal orbital elements $(p,f,g,h,k)$ are prescribed and the terminal true longitude $L(t_f)$ is left free. Consequently, the associated transversality condition becomes $\lambda_L(t_f)=0$~\cite{levineControlSystemsHandbook2018}. This relaxation enlarges the feasible terminal manifold and reduces TPBVP sensitivity to the number of revolutions, eliminating the need to select among solutions with different revolution counts and decreasing the likelihood of continuation path folds and bifurcations~\cite{zhangGTOC11Results2023a,guo_minimum-time_2023}. Figure~\ref{fig:homotopy_bifurcation} illustrates the effect of this relaxation. The only difference between the two continuation runs is the terminal constraint: enforcing $L(t_f)$ (point-to-point) versus $\lambda_L(t_f)=0$ (orbital transfer). The point-to-point continuation exhibits multiple zero crossings of $\det\Phi$, indicating local folds, whereas the orbital-transfer continuation maintains $\det\Phi$ away from zero, demonstrating smoother and more robust convergence.

\begin{figure}[t]
\centering
\includegraphics[width=0.8\linewidth]{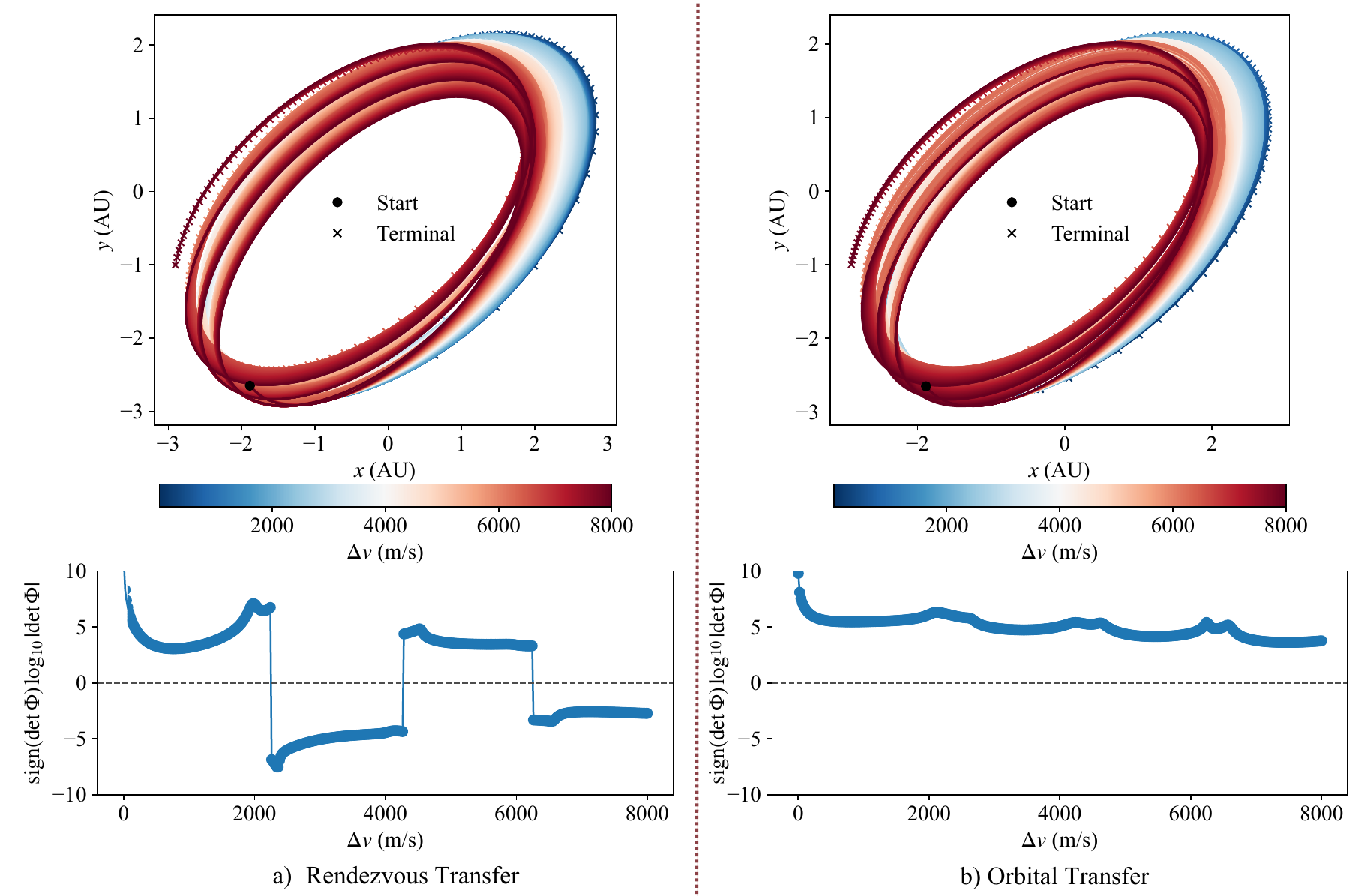}
\caption{Continuation behavior under point-to-point and orbital transfer constraints.}
\label{fig:homotopy_bifurcation}
\end{figure}

Although the above relaxation eliminates the ambiguity, the resulting trajectories are orbital transfers rather than the point-to-point transfers required for network training. Therefore, an additional step is employed. 
After an orbital transfer trajectory is obtained, random segments are extracted from the trajectory. The states at the beginning and end of each segment are then treated as the initial and terminal conditions of a new point-to-point transfer problem, while the extracted segment trajectory itself serves as the corresponding optimal solution. 
The validity of this construction relies on the fact that the extracted sub-trajectories inherit the optimality properties of the original trajectory. A rigorous proof is provided in the supplementary material.

As a result, the extracted samples remain optimal point-to-point transfers and do not reintroduce the label ambiguity associated with multiple-revolution branches.

\subsection{Neural Network Training}
\label{sec:nn_training_multirev}

\begin{figure}[tbh!]
  \centering
  \includegraphics[width=0.8\linewidth]{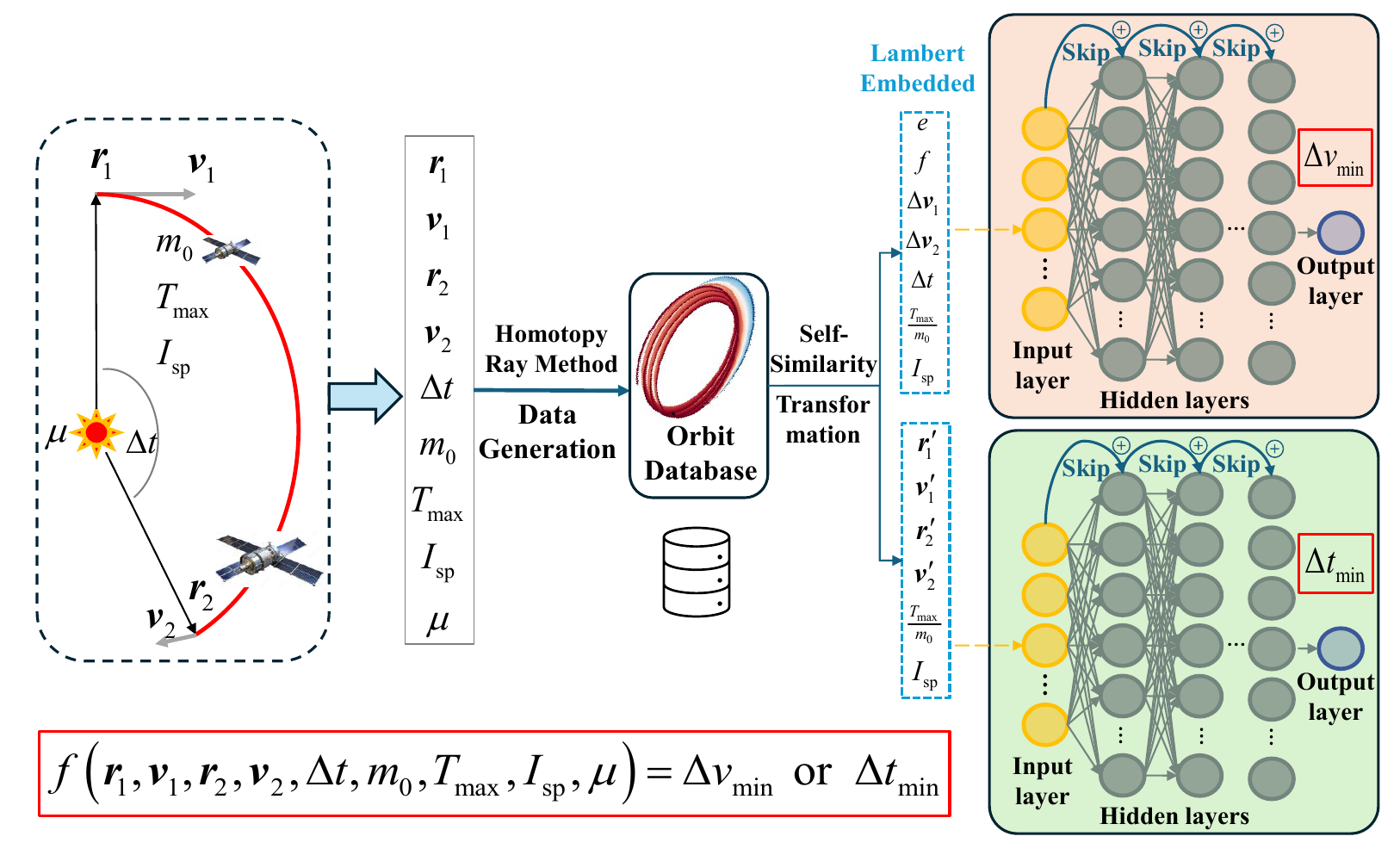}
  \caption{The input and output of the multi-rev model.}
  \label{fig:Figure_Input_Out_Multirev}
\end{figure}%

Figure~\ref{fig:Figure_Input_Out_Multirev} illustrates the final input–output structure of the multi-revolution models.
To capture the increased nonlinearity of multi-revolution transfers—arising from longer arcs, more intricate control switching structures, and greater sensitivity to inputs, we employ a residual multilayer perceptron with skip connections, inspired by deep residual networks~\cite{he2016deep}. The effectiveness of this architecture was validated against a vanilla MLP with identical hyperparameters (9 hidden layers, 128 neurons, GeLU activation, identical learning rate, batch size, and epochs). The residual network reduced the median validation error from 0.3735\% (vanilla MLP) to 0.2824\%, confirming the advantage of skip connections in capturing the complex mappings of multi-revolution transfers.

Systematic hyperparameter tuning was performed on fuel-optimal transfers, exploring the number of hidden layers (HL), neurons per layer (N), and activation functions. Table~\ref{tab:table_results_multirev} summarizes the results in terms of mean absolute error (MAE) and mean absolute percentage error (MAPE) across training, validation, and test sets. Both network depth and width substantially influence performance: increasing from 5 to 9 layers and from 64 to 128 neurons consistently reduces error. Among all configurations, the 9-layer, 128-neuron network with Softplus activation achieves the lowest MAE and MAPE across all datasets. This architecture is also employed for time-optimal transfers.

\begin{table}[h!]
  \small
  \centering
  \caption{Performance of residual network architectures with different settings.}
  \begin{tabular}{ccccccccc}
    \toprule
    HL & N   & Activation & Train MAE              & Train MAPE & Val MAE                & Val MAPE & Test MAE               & Test MAPE \\
    \hline
    5  & 64  & GeLU        & $1.13 \times 10^{-2}$  & 1.41\%     & $1.14 \times 10^{-2}$  & 1.43\%   & $1.14 \times 10^{-2}$  & 1.44\%    \\
    9  & 64  & GeLU        & $9.25 \times 10^{-3}$  & 1.12\%     & $9.59 \times 10^{-3}$  & 1.64\%   & $9.58 \times 10^{-3}$  & 1.20\%    \\
    5  & 128 & GeLU        & $5.42 \times 10^{-3}$  & 0.63\%     & $5.65 \times 10^{-3}$  & 0.67\%   & $5.66 \times 10^{-3}$  & 0.68\%    \\
    9  & 128 & GeLU        & $3.53 \times 10^{-3}$  & 0.40\%     & $3.85 \times 10^{-3}$  & 0.45\%   & $3.85 \times 10^{-3}$  & 0.45\%    \\
    9  & 128 & ReLU        & $5.00 \times 10^{-3}$  & 0.57\%     & $5.44 \times 10^{-3}$  & 0.63\%   & $5.43 \times 10^{-3}$  & 0.64\%    \\
    9  & 128 & SiLU        & $3.13 \times 10^{-3}$  & 0.36\%     & $3.39 \times 10^{-3}$  & 0.39\%   & $3.39 \times 10^{-3}$  & 0.40\%    \\
    9  & 128 & Softplus    & $2.99 \times 10^{-3}$  & 0.34\%     & $3.23 \times 10^{-3}$  & 0.38\%   & $3.22 \times 10^{-3}$  & 0.38\%    \\
    \bottomrule
  \end{tabular}
  \label{tab:table_results_multirev}
\end{table}

\subsection{Perfomance Assessment}
\label{sec:results_multirev}

\begin{figure}[tbh!]
  \centering
  \includegraphics[width=0.4\linewidth]{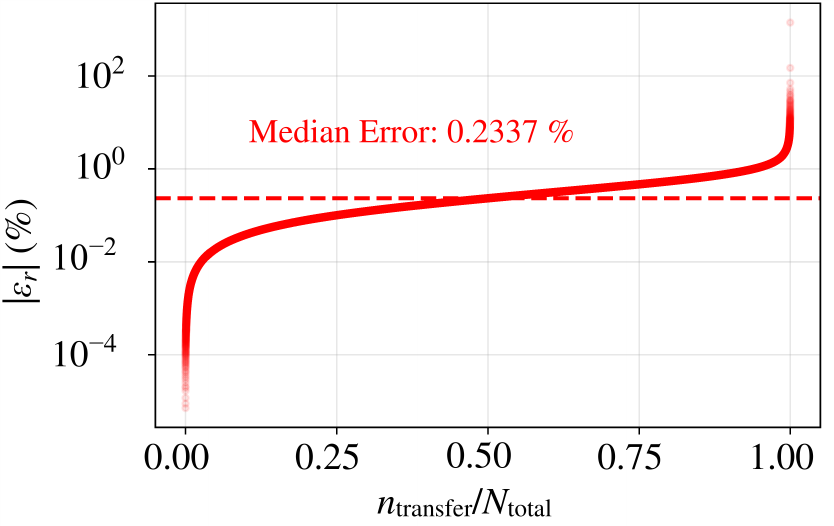}
  \hfill
  \includegraphics[width=0.4\linewidth]{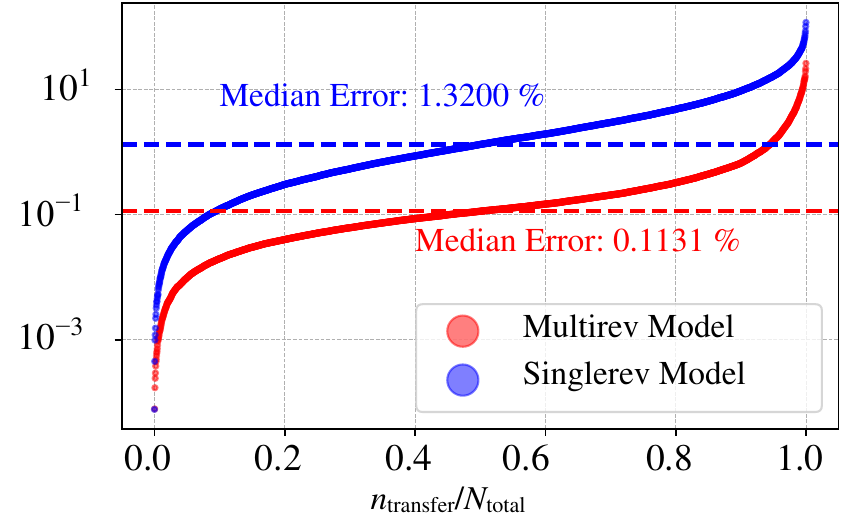}
  \caption{Performance of single and multirev $\Delta v$ models.}
  \label{fig:sorted_rel_error_multirev}
\end{figure}

Figure~\ref{fig:sorted_rel_error_multirev} presents the sorted relative prediction errors on the test set for the optimal Softplus architecture. The left panel shows the multi-revolution model, achieving a median error of 0.2337\%, demonstrating robust approximation across the full range of multi-revolution transfers. The right panel compares the multi-revolution model against the single-revolution model evaluated on the single-revolution test set. The multi-revolution network outperforms the single-revolution model, suggesting that exposure to higher-revolution data improves generalization even for simpler, zero-revolution transfers.

\begin{figure}[tbh!]
  \centering
  \includegraphics[width=0.45\linewidth]{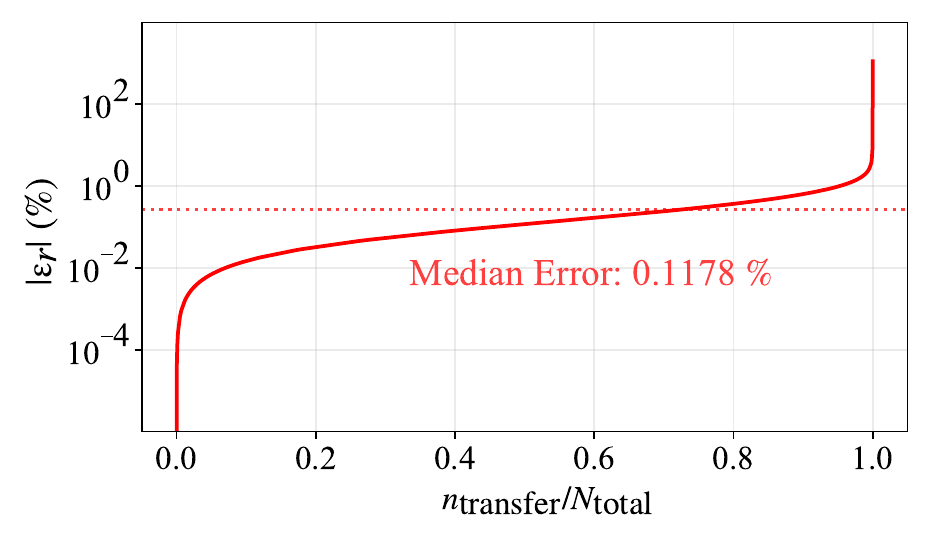}
  \caption{Performance of multirev $\Delta t$ models.}
  \label{fig:sorted_rel_error_multirev_tmin}
\end{figure}%

\begin{figure}[tbh!]
\centering
\includegraphics[width=0.6\linewidth]{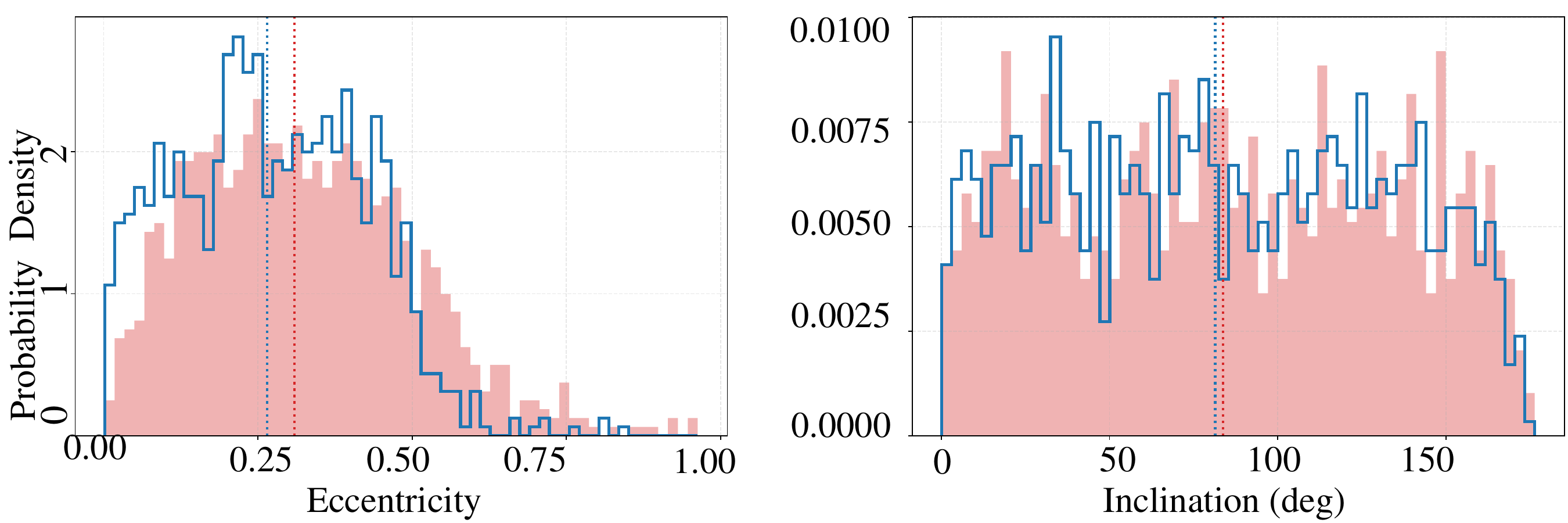}\
\includegraphics[width=0.6\linewidth]{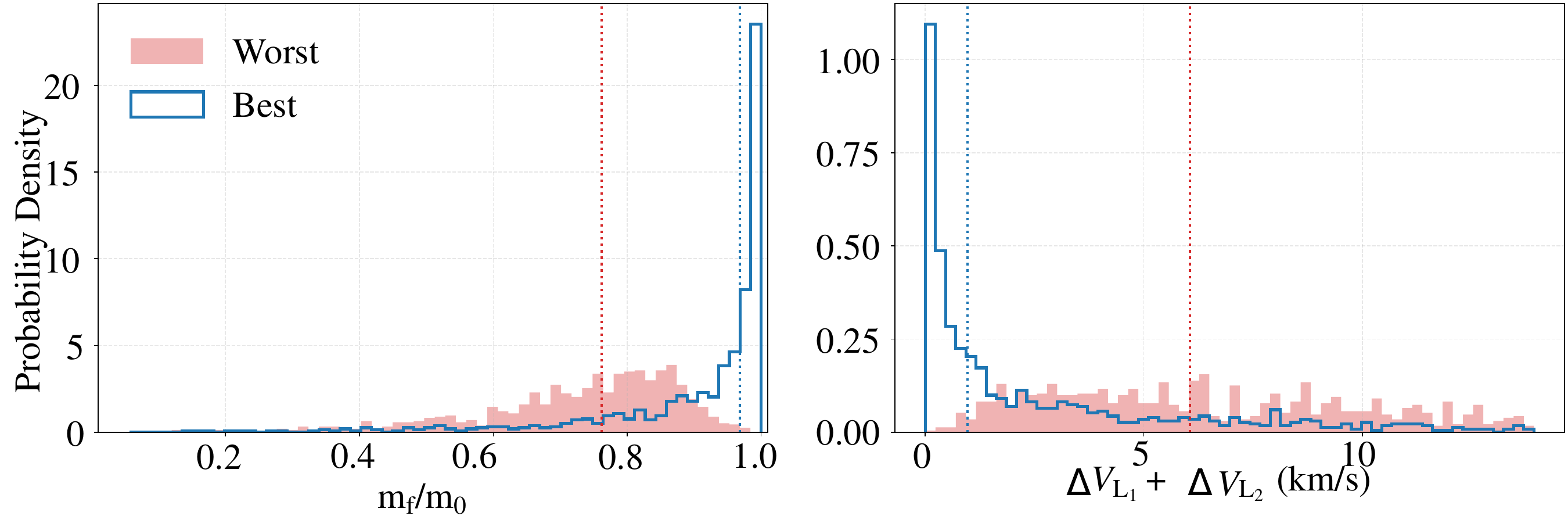}\
\includegraphics[width=0.6\linewidth]{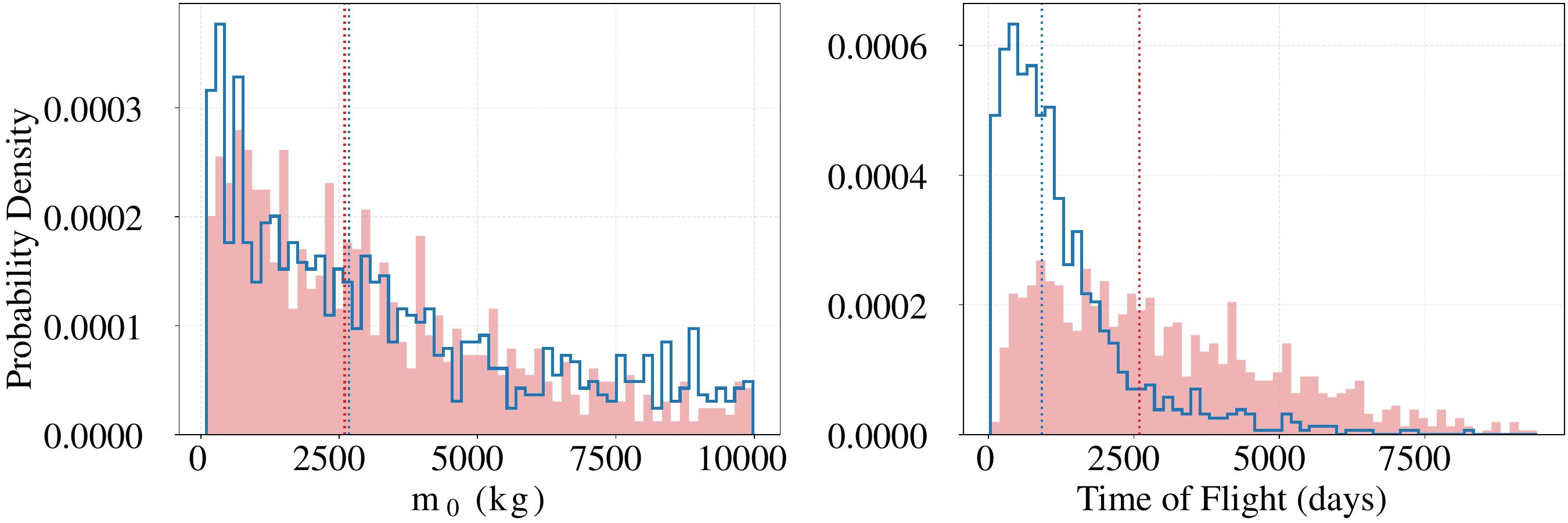}
\caption{Comparison of transfer characteristics associated with best- and worst-performing predictions.}
\label{fig:statistics_best_worst_multirev_model}
\end{figure}

For time-optimal transfers, the multi-revolution model achieves comparable performance, as shown in Figure~\ref{fig:sorted_rel_error_multirev_tmin}.
To investigate factors affecting prediction accuracy, we analyzed the top and bottom 1,000 test cases. Figure~\ref{fig:statistics_best_worst_multirev_model} reports distributions of key transfer features. Errors are largely insensitive to eccentricity, inclination, and initial mass. However, more challenging transfers, characterized by higher fuel consumption, longer time-of-flight, and larger Lambert cost, tend to appear in the worst-performing cases. The median final-to-initial mass ratio of the worst 1,000 cases is roughly 30\% lower than that of the best 1,000 cases.

The distribution of revolution counts further highlights model behavior. Among the top 1,000 cases, 742 are zero-revolution, 158 one-revolution, 79 two-revolution, and 21 three-revolution transfers. In contrast, the worst 1,000 cases include 172 zero-revolution, 334 one-revolution, 325 two-revolution, and 169 three-revolution transfers. This trend indicates that accuracy is highest for zero-revolution trajectories, while performance decreases with increasing number of revolutions due to the higher nonlinearity and control complexity.

\section{Results}
\label{sec:results}

This section presents the final performance of the proposed neural network models and evaluates their applicability through three complementary tasks: (i) validation on public datasets, (ii) application to the GTOC4 multi-flyby asteroid trajectory design problem, and (iii) porkchop plot generation for asteroid rendezvous analysis. The goal is to assess the models’ generalization capability, practical usability in mission design, and computational efficiency.

Validation on public datasets tests the models under varying data distributions and verifies the effectiveness of the proposed homotopy ray data generation methodology. The GTOC4 multi-flyby design task provides a benchmark for sequential mission planning, evaluating the models’ accuracy and integration within trajectory optimization frameworks. Finally, the porkchop plot generation task highlights the computational efficiency and usability of the models compared to traditional optimal control methods.

\subsection{Training Result}
\label{sec:training_result}

The final models were trained on an extended dataset generated via the proposed homotopy ray procedure, comprising up to approximately 100 million samples. 
The performance of the fuel-optimal model is summarized in Table~\ref{tab:dv_model_final_performance}. The evaluation metric is the final mass $m_f$, which reflects the cumulative fuel consumption of the transfer. Both single- and multi-revolution models outperform previously reported methods in terms of mean absolute error (MAE) and mean relative error (MRE). Specifically, the models achieve average relative errors of 0.014\% for single-revolution and 0.003\% for multi-revolution transfers. The corresponding MAE in $\Delta v_{\min}$ is approximately 3.38~m/s (single-revolution) and 99.40~m/s (multi-revolution).
Table~\ref{tab:dt_model_final_performance} reports the predictive accuracy of the time-optimal model. Average relative errors in $\Delta t_{\min}$ are 0.54\% for multi-revolution and 0.63\% for single-revolution transfers, demonstrating the models’ strong capability in evaluating transfer feasibility and time-optimal performance.

\begin{table}[htbp]
  \centering
  \caption{$\Delta v$ Model Performance Compared to Existing Models}
  \begin{threeparttable}
    \begin{tabular}{lccccc}
      \toprule
                                                         & \multicolumn{3}{c}{ Other studies}        & \multicolumn{2}{c}{ This paper}                                                                                                                                        \\
      \cmidrule(lr){2-4} \cmidrule(lr){5-6}    Parameter & Zhu~\cite{zhuFastEvaluationLowThrust2019} & Li~\cite{liDeepNetworksApproximators2020} & Acciarini~\cite{acciariniComputingLowthrustTransfers2024a} & Singlerev Model               & Multirev model                \\
      \midrule
      $a$, AU                                            & $2.0\sim3.0$                              & $2.0\sim3.5$                              & $0.9\sim4.0$                                               & any                           & any                           \\
      $e$                                                & $0\sim0.4$                                & $0\sim0.1$                                & $0\sim0.48$                                                & $0\sim1.0$                    & $0\sim $ 1.0                  \\ 
      $i$, deg                                           & $0\sim20$                                 & $0\sim10$                                 & $0\sim30$                                                  & any                           & any                           \\
      $\Delta t,n$                                       & $0\sim0.48$                               & $0\sim0.3$                                & $0\sim1.99$                                                & $0\sim0.99$                   & $0\sim3.99$                   \\
      $a_s$ (min), m/s$^2$                               & $1.5\times10^{-4}$                        & $1.5\times10^{-4}$                        & $7.5\times10^{-5}$                                         & $2.5\times10^{-6}$\tnote{[a]} & $2.5\times10^{-6}$\tnote{[a]} \\
      $a_s$ (max), m/s$^2$                               & $3.8\times10^{-4}$                        & $3.0\times10^{-4}$                        & $8.6\times10^{-4}$                                         & $1.2\times10^{-2}$\tnote{[a]} & $1.2\times10^{-2}$\tnote{[a]} \\
      $I_{\rm sp}$, s                                    & 3000                                      & 3000                                      & 4000                                                       & $700\sim9000$\tnote{[a]}      & $700\sim9000$\tnote{[a]}      \\
      \midrule
      $m_f$ MAE, kg                                      & /                                         & /                                         & 97.48\tnote{[b]}                                           & 0.42                          & 13.34                         \\
      $m_f$ MRE, \%                                      & 0.37\%  \tnote{[c]}                                  & 0.50\%  \tnote{[c]}                                  & 2.68\%\tnote{[b]}                                          & 0.014\%                       & 0.003\%                       \\
      \bottomrule
    \end{tabular}%
    \begin{tablenotes}
      \footnotesize
      \item[a] The value is calculated at 1 AU.
      \item[b] The value is calculated by first 1 million data.
      \item[c] The value is taken from the corresponding paper.
    \end{tablenotes}
  \end{threeparttable}
  \label{tab:dv_model_final_performance}%
\end{table}%

It should be noted that the studies summarized in Table~\ref{tab:dv_model_final_performance} were developed using datasets of different sizes, parameter ranges, and train/validation/test partitions.  Furthermore, the underlying data distributions are not necessarily similar and comparable across studies. Consequently, the comparison should be interpreted as illustrative of overall model applicability and performance across some specific input distributions and bounds, rather than as a strict one-to-one benchmark.
For Zhu~\cite{zhuFastEvaluationLowThrust2019} and Li~\cite{liDeepNetworksApproximators2020}, no public datasets are released; therefore, we can only report and compare the domain bounds explicitly stated in their papers, while the underlying data distributions (and any train/test sampling differences) remain unknown. 
As a result, a strictly fair matched-domain, dataset-level evaluation across these studies cannot be rigorously defined. 
In contrast, Acciarini et al.~\cite{acciariniComputingLowthrustTransfers2024a} provides an openly accessible dataset, which enables a more direct comparison.
Accordingly, in the next subsection~\ref{sec:thirdparty} we include an additional comparison on the dataset from~\cite{acciariniComputingLowthrustTransfers2024a}.

\begin{table}[htbp]
  \centering
  \caption{$\Delta t$ Model Performance Comparison}
  \begin{threeparttable}
    \begin{tabular}{lccc}
      \toprule
                               & \multicolumn{1}{c}{Other studies}        & \multicolumn{2}{c}{This paper}                                 \\
      \cmidrule(lr){2-2} \cmidrule(lr){3-4}
      Parameter                & Guo~\cite{guoDNNEstimationLowthrust2023} & Singlerev Model                & Multirev Model                \\
      \midrule
      $a$, AU                  & $2.0 \sim 3.0$                           & any                            & any                           \\
      $e$                      & $0 \sim 0.4$                             & $0\sim1.0$                     & $0\sim1.0$                    \\
      $i$, deg                 & $0 \sim 20$                              & any                            & any                           \\
      $\Delta t,n$             & $0 \sim 1.35$                            & $0\sim0.99$                    & $0\sim 3.99$                  \\
      $a_s$ (min), m/s$^2$     & $1.2\times10^{-4}$                       & $2.5\times10^{-6}$\tnote{[a]}  & $2.5\times10^{-6}$\tnote{[a]} \\
      $a_s$ (max), m/s$^2$     & $6.0\times10^{-4}$                       & $1.2\times10^{-2}$\tnote{[a]}  & $1.2\times10^{-2}$\tnote{[a]} \\
      $I_{\rm sp}$, s          & 3000                                     & $700\sim9000$\tnote{[a]}       & $700\sim9000$\tnote{[a]}      \\
      \midrule
      $\Delta t_{\min}$ MAE, days & /                                        & 2.56                           & 11.24\\ 
      $\Delta t_{\min}$ MRE, \%   & 1.30\%\tnote{[b]}                        & 0.63\%                         & 0.54\% \\
      \bottomrule
    \end{tabular}%
    \begin{tablenotes}
      \footnotesize
      \item[a] The value is calculated at 1 AU.
      \item[b] Best reported MRE for fast transfers~\cite{guoDNNEstimationLowthrust2023} (dataset augmented with $m=3$).
    \end{tablenotes}
  \end{threeparttable}
  \label{tab:dt_model_final_performance}%
\end{table}

Beyond accuracy, the results emphasize the versatility of the proposed framework. Unlike previous models that impose constraints on specific orbital elements (e.g., semi-major axis $a$, eccentricity $e$, inclination $i$), the present model accommodates arbitrary celestial bodies. Propulsion parameters, including acceleration $a_s$ and specific impulse $I_{\rm sp}$, span a wide operational range, covering nearly all practically relevant configurations. This generality enables straightforward application across diverse mission scenarios, without the need for retraining or additional data generation, offering mission designers and planetary scientists a flexible and efficient tool for preliminary trajectory assessment.

\subsection{External Dataset Validation}
\label{sec:thirdparty}

As an independent test, we evaluate the proposed model on a public dataset, which contains one million low-thrust transfer samples and represents the only publicly available dataset in this domain~\cite{acciariniComputingLowthrustTransfers2024a}. In the publication that accompnies the dataset release, the authors include prediction results on the dataset, which enables a more direct comparison. Evaluation metrics are consistent with previous sections, using MAE and MRE.

\begin{figure}[hbt!]
\centering
\includegraphics[width=0.55\linewidth]{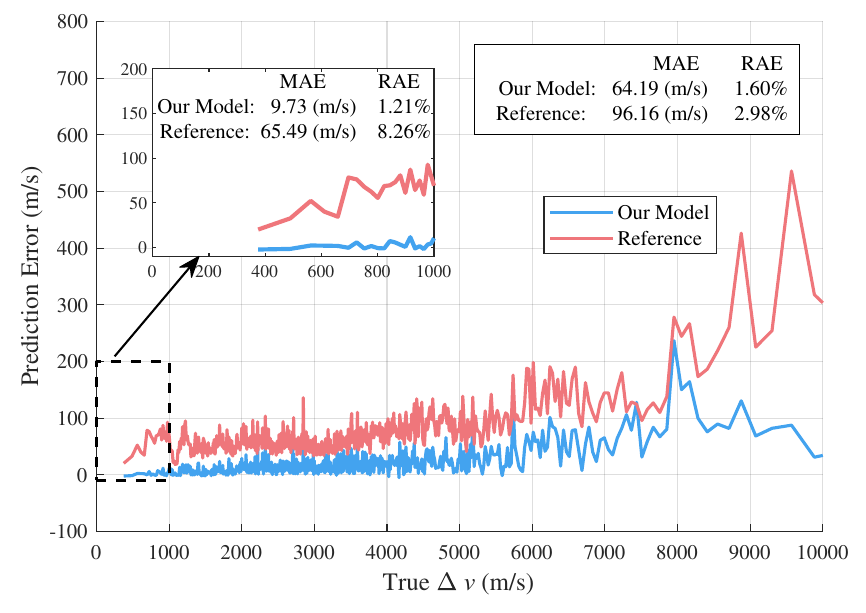}
\caption{Validation of the public dataset.}
\label{fig:3rdparty}
\end{figure}

Figure~\ref{fig:3rdparty} presents the performance of the $\Delta v$ model across the dataset. Ground-truth $\Delta v$ values were sorted in ascending order and divided into consecutive windows of 1,000 samples each. For every window, the average $\Delta v$ and corresponding MAE were computed to visualize error trends.

Analysis indicates that the dataset of \cite{acciariniComputingLowthrustTransfers2024a} contains relatively few transfers with $\Delta v < 200$~m/s, as their data generation strategy starts from time-optimal solutions and progressively transitions toward fuel-optimal cases. Consequently, their model achieves higher accuracy in the high $\Delta v$ regime ($\Delta v > 1000$~m/s), while exhibiting reduced reliability in the low-thrust region. In contrast, the proposed model maintains consistently high accuracy across the entire $\Delta v$ spectrum, highlighting the effectiveness of the homotopy ray-based data generation strategy described in Sec.~\ref{sec:DataGeneration}.

Our proposed model is trained on an extended dataset generated via the homotopy ray procedure, comprising more than 100 times the number of samples used in \cite{acciariniComputingLowthrustTransfers2024a} training. This substantial increase in dataset size, together with broader coverage of specific impulse, inclination, eccentricity, and thrust acceleration ranges, should be taken into account when interpreting the performance differences. These findings are consistent with the trend discussed earlier in this work, suggesting that increasing dataset scale and diversity leads to improved predictive accuracy, and indicating a clear scaling effect in learning fuel-optimal low-thrust transfer behavior.

\subsection{Multi-Flyby Asteroid Mission Design}
\label{sec:multi-flyby}

To demonstrate practical applicability, the proposed models are evaluated on the GTOC4 multi-flyby asteroid trajectory design problem~\cite{Grigoriev2013}. This benchmark involves a ten-year mission with multiple asteroid flybys, culminating in a final rendezvous.
Two comparative results are considered:
Result 1: Segment-wise predictions based on the state-of-the-art GTOC4 trajectory from the University of Jena, involving flybys of 49 asteroids. The proposed model predicts the required velocity increments using the extracted flyby segments; Result 2: Neural network-based trajectory optimization that adjusts flyby timing and relative velocities to produce a more fuel-efficient mission. Details of this optimization procedure are outside the scope of this work and are reported separately~\cite{zhang2025globaloptimalitymultiflybyasteroid}.

\begin{figure}[hbt!]
  \centering
  \includegraphics[width=0.7\linewidth]{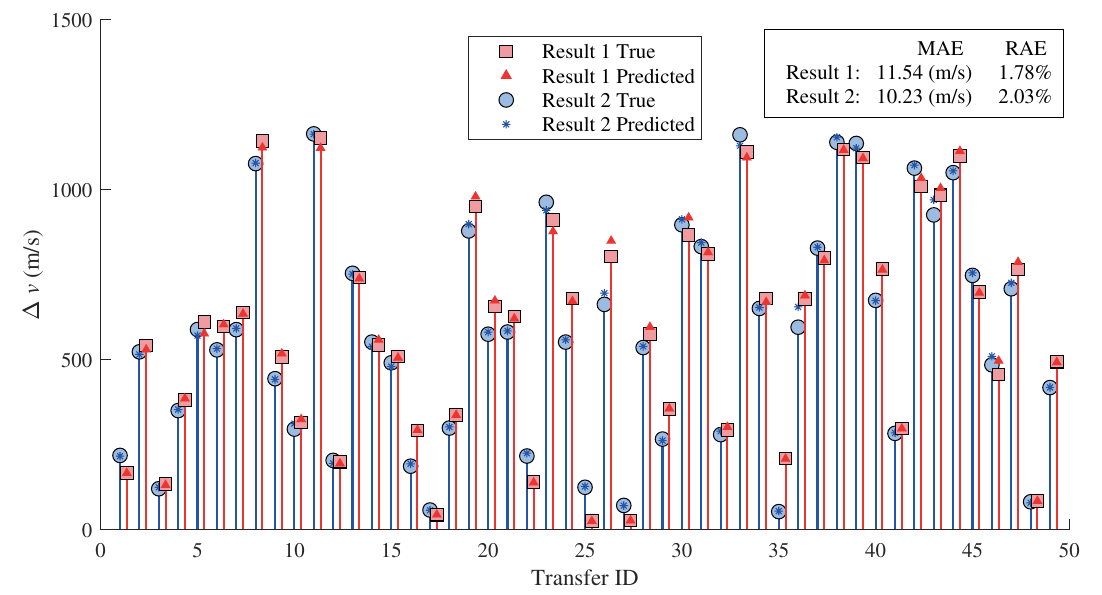}
  \caption{Validation of the GTOC4 problem results.}
  \label{fig:GTOC4}
\end{figure}

Figure~\ref{fig:GTOC4} compares the predicted versus actual $\Delta v$ values across all flyby segments. The model exhibits strong predictive performance, with accurate estimates even for segments requiring less than 100~m/s of velocity increment. This highlights the importance of including low-thrust trajectories in the training dataset to ensure reliability in precision mission design.

\subsection{Mission Analysis}
\label{sec:porkchop}
This subsection demonstrates a porkchop-plot workflow for rapid preliminary screening of low-thrust Earth–asteroid rendezvous missions using the proposed surrogate models. The benchmark scenario considers a CubeSat departing from Earth under a launch-energy constraint and rendezvousing with a near-Earth asteroid. The propulsion is assumed constant at 1~AU, with maximum thrust $T_{\max}=\SI{1.82}{mN}$ and specific impulse $I_{\rm sp}=\SI{3600}{s}$. The launcher provides a maximum Earth-departure hyperbolic excess speed of $v_{\infty,\max}=\SI{4}{km/s}$, with unconstrained departure direction.

The motivation is large-scale target screening. Even after applying maneuverability-based filters, hundreds to thousands of asteroid candidates remain from a population exceeding one million objects. For low-thrust rendezvous, each grid point of a conventional porkchop plot requires solving a numerical optimal-control problem. This becomes computationally prohibitive for dense grids, multiple targets, or iterative preliminary design analyses, particularly when propulsion parameters or mission constraints vary.

A porkchop plot is constructed by sweeping the departure epoch $t_0$ and time of flight $\Delta t$ on a prescribed grid. Since the launch-provided departure velocity is not uniquely defined, the Earth-departure hyperbolic excess velocity $\pmb{v}{\infty}$ is treated as a decision variable at each grid point, constrained by $|\pmb{v}\infty|2 \le v{\infty,\max}$. Particle swarm optimization is employed to select $\pmb{v}\infty$ that minimizes the surrogate-predicted cost. At each grid point, the $\Delta v$ surrogate estimates $\Delta v{\min}(t_0,\Delta t,\pmb{v}\infty)$, while the $\Delta t$ surrogate provides the minimum feasible time $\Delta t{\min}(t_0,\pmb{v}_\infty)$ to complete the transfer. Feasibility is enforced via a penalty for time-of-flight violations.

\begin{figure}[hbt!]
  \centering
  \includegraphics[width=0.8\linewidth]{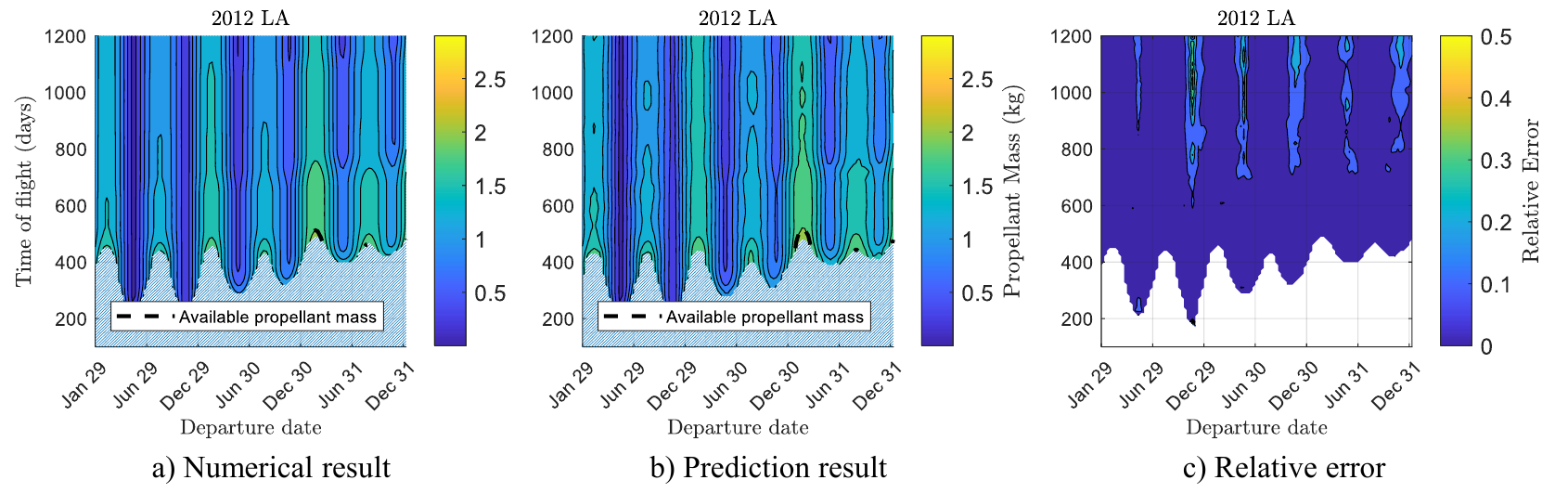}
  \caption{Porkchop plot for the 2012 LA Asteroid.}
  \label{fig:porkchop_2012LA}
\end{figure}

\begin{figure}[hbt!]
  \centering
  \includegraphics[width=0.8\linewidth]{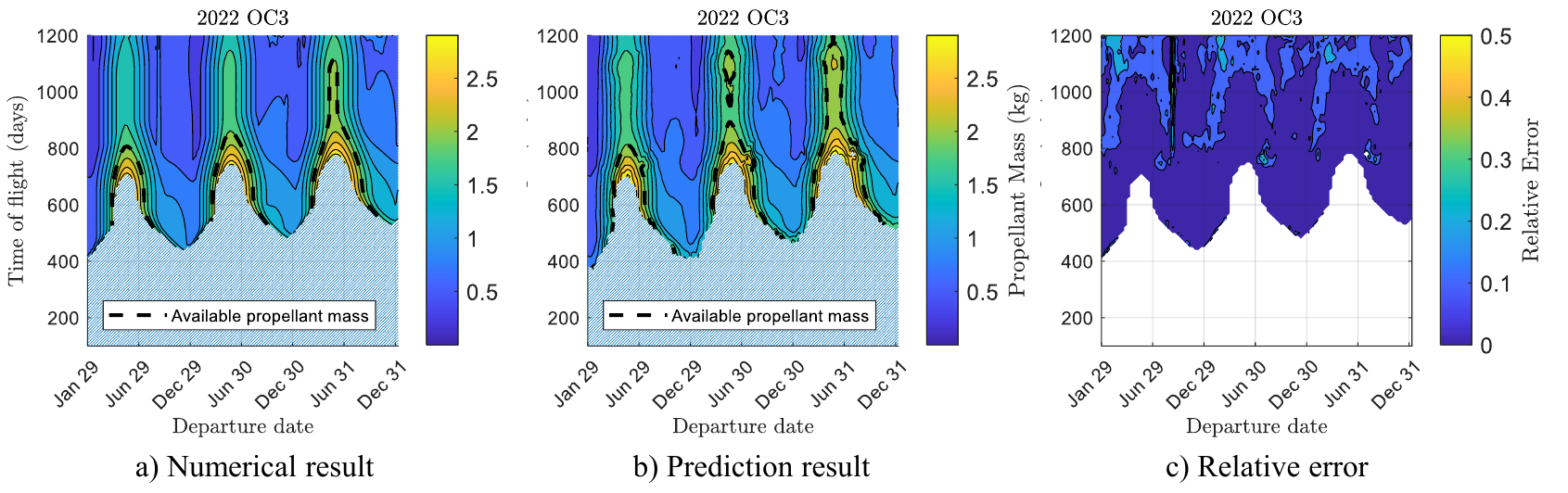}
  \caption{Porkchop plot for the 2022 OC3 Asteroid.}
  \label{fig:porkchop_2022OC3}
\end{figure}

Three representative target asteroids are considered: 2012~LA and 2022~OC3 (Figs.~\ref{fig:porkchop_2012LA} and \ref{fig:porkchop_2022OC3}). For each target, surrogate-generated porkchop plots are compared with reference plots computed using a high-fidelity optimal-control solver on the same grid. For 2012~LA, the surrogate accurately reproduces the cost landscape, including low-$\Delta v$ windows and feasibility boundaries, with relative errors typically below 10\% across most of the feasible domain.Figure~\ref{fig:porkchop_2022OC3} highlights a limitation of traditional optimal-control methods, where solution-continuation techniques may fail to converge in the presence of multiple local optima. In contrast, the surrogate-based approach, coupled with global optimization, reliably identifies feasible solutions, demonstrating robustness and practical effectiveness. Moreover, while high-fidelity optimal-control solvers typically require days to generate a single porkchop plot, the proposed surrogate model computes results in negligible time, enabling rapid preliminary mission analysis.

\section{Conclusion}
\label{sec:conclusion}

This paper has presented a neural network–based framework for rapid and accurate estimation of fuel consumption and transfer feasibility in low-thrust trajectory design. The proposed models advance the performance by enabling broad applicability across a wide variety of mission scenarios without the need for retraining or task-specific dataset generation. The accuracy, generalization, and computational efficiency of the proposed models were extensively validated. This capability stems from two complementary innovations:
At the data level, we observed an approximate empirical scaling law for low-thrust trajectories, which guided the generation of a large-scale dataset encompassing over 100 million trajectory samples by the homotopy ray method. The dataset spans single- and multi-revolution transfers, a wide range of orbital elements, thrust levels, and specific impulses, as well as diverse mission geometries.

At the astrodynamics level, inherent symmetries in the fuel- and time-optimal low-thrust control problem were exploited to reduce input dimensionality. Rotational invariance, non-dimensionalization, and Lambert solver solutions were incorporated into the input pipeline, providing the neural networks with physically meaningful and compact representations of mission states. 

In summary, this work reduces the computational and engineering overhead in mission designs and lowers the barrier imposed by optimal-control expertise. This enables even non-specialists, such as planetary scientists, to perform rapid reachability and cost assessments 
for next-generation interplanetary or near-Earth mission studies and to iterate mission concepts more efficiently.

\section*{Acknowledgments}
This work was supported in part by the European Union's Horizon Europe Program for Marie Skłodowska-Curie Actions under Grant 101263692.
The authors acknowledge ChatGPT for English language polishing.

\section*{Codes and Data Availability Statement}
The model used in this paper is available on GitHub at \url{https://github.com/zhong-zh15/neural-low-thrust-approximator}, and supports platforms including C++, Python, and MATLAB. Two different versions are provided for both the C++ and Python implementations.
The datasets associated with this paper are available from the open-access
repository Zenodo at:  \url{https://doi.org/10.5281/zenodo.18769170}.

\bibliography{references}

\end{document}